\documentclass[fleqn,10pt]{article}
\usepackage[utf8]{inputenc}
\usepackage[T1]{fontenc}
\usepackage{
amsmath,% AMS basic math stuff
amsthm,% AMS theorem defining stuff
amsfonts,% defines the blackboard bold fonts for \Z, \R, etc
array,
authblk,
booktabs,
comment, % commenting
dirtytalk,
float,
geometry, 
graphicx,
hyperref,
enumitem, % item formatting
longtable,% used to create the tcproof environment below
verbatim,% allows for verbatim output, and also covering up stuff in comments
xspace,% adds an extra space an the end of some commands
multicol,% allows multicolumn output
multirow, % allows multirow output
natbib,
subfig,
tabu,
tikz,% creates the tikzpicture drawing environment
tkz-graph, %additional tikz package
framed,% used to color in the TIscreen environment below
}
\newcommand{\A}{\mathcal{A}}
\newcommand{\D}{\mathcal{D}}
\newcommand{\G}{\mathcal{G}}

\newcommand{\specialcell}[2][c]{\begin{tabular}[#1]{@{}c@{}}#2\end{tabular}}
\newcommand{\keywords}[1]{\textbf{\textit{Keywords---}} #1}
\newcolumntype{P}[1]{>{\centering\arraybackslash}p{#1}}
\newcolumntype{M}[1]{>{\centering\arraybackslash}m{#1}}
\newtheorem{lemma}{Lemma}[section]

\theoremstyle{definition}
\newtheorem{definition}[lemma]{Definition}
\newtheorem{example}{Example}[section]
\newenvironment{comments}{}{}

\newcommand{\myrule}{\nobreak\par\noindent\mbox{}\hspace*{.333333\textwidth}%
            \rule{.333333\textwidth}{.01in}\hspace*{.33333\textwidth}\par}
\makeatletter
  \g@addto@macro\enddefinition\myrule
  \g@addto@macro\endexample\myrule
  \g@addto@macro\endcomments\myrule
\makeatother

\title{A Formal Critique of the Value of the Colombian Páramo}
\author{Juan Afanador}
\affil{University of Edinburgh}
\date{27 April 2020}

\begin{document}

\maketitle

\begin{abstract}
    This article presents conceptual and methodological frameworks to prioritise interventions on the Colombian Páramo. The mode of analysis that our work takes up is that of questioning value and related categories as definite empirically perceived phenomena. We contend that the valuation of ecosystem services --- even in its post-normal forms --- and the ecosystem services framework not only fail to examine value-based categories, but reproduce the problematic aspects of value-based social relations, which ultimately bear on the ecological issues affecting the Páramo. Upon this premise we set out to formalise a (computational) dialogical scenario where arguments stating distinct, and often contradictory, actions delineate possible forms of appropriating the Páramo, while motivating the examination of their defining sociality.  
\end{abstract}{}

\section{Introduction}
\label{sec:intro}

The ecosystem services framework (ESF) renders the idea of people benefiting from their natural environment amenable to the accrual of economic profit or, more generally, value \citep{degroot2002,assessment2005ecosystems, carpenter2006}. ESF thus beckons the valuation of ecosystem services (VES) as a means to signalling nature's contribution to the (re)production of value \citep{barbier2009valuation,villa2009aries,fisher2010valuing,gomez2016concepts}; for value is the central category of modern capitalist societies, and the valorisation of value --- i.e., economic growth sublimated into economic development --- their driving force (see, e.g., \cite{mankiw2016principles} and \cite{holden2017imperatives}). VES is, in this sense, inscribed in an interpretive approach to modern capitalist praxis, not only invoking assumptions that are instrumentally validated in a retroactive manner, but also taking for granted precisely those historical and material conditions which VES is meant to interpret and, in doing so, reproduce. 

Overlooking the historical basis of ESF and VES has important practical consequences. When VES practitioners elicit value, a moment or specific field of the social praxis embodied in the valorisation of value is inaugurated, allowing value to mediate other social constructs built around the idea of nature. Since the patterns of actions that make up the capitalist social praxis are presupposed within this new ambit, value takes on a transhistorical quality that justifies its all-encompassing and unreflective usage (see, e.g., \cite{badura2016valuing} and \cite{gomez2015ecological}). In this manner, axiological considerations --- e.g., the belief in nature's intrinsic goodness (see, e.g., \cite{davidson2013relation}) --- and more concrete acts on nature --- e.g. the importance of implementing agripastoral practices compatible with environmental conservation (see, e.g., \cite{lliso2020payments}) --- are \textit{naturally} linked to value, prices and their accompanying categories --- in this manner, modern capitalist praxis is reproduced within new and existing domains.

Given the neoclassical antecedents of VES --- that is, given the econometric origins of its methods and their various degrees of microeconomic soundness --- it may be tempting to advance the idea that non-neoclassical approaches to VES may prevent the above moment of social praxis (see, e.g., \cite{gomez2016concepts} and \cite{kumar2008valuation}). However, the sole invocation of the valorisation of value inherent in valuation --- in any approach to valuation --- induces a form of fetishistic use of value, i.e., the implicit subjectification of (the category of) value as an agent of social interaction --- as an ``automatic subject", as it were \citep{kurz2019substance}. Therefore, a similar process of social reproduction is prompted by non-neoclassical approaches to VES --- e.g., those employing non-monetary techniques under the principles of integral valuation (see, e.g., \cite{jacobs2016new} and  \cite{gomez2014state}) --- so long as valuation is constituted as an affirmative action around the category of value.

Out of these observations, our work is laid out as a critique of VES. Here we present an attempt to break with the basic categories of VES, in the context of the social constructs that we identified with the Guantiva-La Rusia Páramo in Colombia. This break alludes to the analytical negation of value's underlying (formal) logic, which primarily rests upon the Enlightenment ideology, classical logics, and the view that consistency imposes a bound on rationality.

Our main argument affirms that the relations of value-dissociation, i.e., the fetishistic use of value and its gender dissociation manifestations \citep{scholz2013}, have long become the \say{natural} basis of social interaction, through a process of assimilation and reproduction not unlike the one described before \citep{noys2010persistence, hemmens2019critique}. We maintain that within this process, VES acts as a form of theoretical praxis furthering value fetishism towards the constitution of the valorisation of value as an inexorable mandate to which everyone within its orbit must conform under the penalty of (economic) ostracism. We trace the root of this assimilation process to the making of a consistent world, i.e., one that does not contradict capitalist social praxis, by means of a narrative of the conservation and protection of nature through its rational exploitation (see, e.g., \cite{assessment2005ecosystems} and \cite{lliso2020payments}). Hence, we examine the rationality behind the decisions made in a community-based organization of local farmers embracing the relations of value-dissociation within the Guantiva-La Rusia Páramo, employing a paraconsistent logical model of discursive analysis that reveals the immanent contradictions and material implications of the conscious reshaping of their sociality. 

Embedded in a computational environment where individual arguments interact in a common dialogical setting, our paraconsistent model will not only serve to inquiry the rationality of modern capitalist relations, but will also suggest various actions on the Páramo. In this sense, our work can be seen as a form of counter-valuation. We start off from the category of value and the ESF, and by negating them a distinct methodological praxis can be discerned.

The outcomes of our approach will be presented using graph-theoretic tools, such as network-based arrangements and various centrality measures \citep{newman2001fast, kamvar2003eigentrust, brandes2007centrality, opsahl2010node}. Additional sensitivity analysis will be conducted using Bayesian methods --- Markov chain Monte Carlo (MCMC) methods --- to obtain stable states of the networks  \citep{kruschke2013bayesian}. These statistical and graph-theoretic techniques operate on a meta-analytical level with respect to our logic-based proposal; that is to say, that these latter techniques are but instrumental to conveying our contribution.

The remainder of this paper is structured as follows. Section \ref{sec:rationality} presents a discussion on value, rationality, and VES, which problematises our central theme. Section \ref{sec:paramo} characterises the Páramo from a critical perspective. In Section \ref{sec:logics_contradiction} we present our dialogical model through an example based on classical logic, giving way to a formal rendition of the ideas presented thus far in a Dialectical Dialogical Game in Section \ref{sec:example}. The results of implementing our approach in the context of the Páramo are summarised in Section \ref{sec:results}. The last section discusses our findings and provide suggestions for future work.

\section{Reason and Rationality in Valuing Ecosystem Services}
\label{sec:rationality}
Modern capitalist societies function upon a mode of (re)production rather than a system of circulation \citep{holden2017imperatives, varian2018artificial}. That is, commodities have to be the object of production before they can become the object of circulation; and the interplay of all forms of social mediation occurring therein constitute our mode of life --- the capitalist social praxis \citep{andrews2002commodity}. In this sense, we maintain that value is formed through the expenditure of effort in the production of commodities abstracted into working hours, which is then realised transactionally in its price-regulated money form \citep{kurz2019substance}. Put differently, value is the central category of the capitalist social praxis, for it is the body of abstract labour and exchange, hence inextricably bound up with prices and money.

As argued in Section \ref{sec:intro} the dynamics of value have a rare quality. Since value is manifested as pure social mediation, despite having well-defined historical and material bases, its dynamics are those of an automatic subject \citep{hemmens2019critique,kurz2019substance}. Value is reproduced for the sake of its own reproduction, while conditioning the continuation of the reproduction itself. The subjectification of value is then sealed by the internalisation of these categories as second nature, through our being socialised within them --- a form of social mediation that we term \textit{value fetishism}, in line with the less comprehensive concept of \textit{commodity fetishism} \citep{marx2019capital}.

In general, other usages of the word "value", referring to the assignment of relative importance or the generation of axiological judgements, may, or may not, be confined to such etymology. Within VES' domain, the notion of value is used indistinctly, either adducing an innocuous semantic liberty or invoking an use/exchange interpretation \citep{kosoy2010payments}. The former is but the evocation of said second nature, while the other propounds the essentialisation of the appropriation of nature as a socially productive activity akin to labour ---to abstract labour as the kernel of value (see, e.g., \cite{jacobs2016new}). Both meanings are affirmed by the ESF through its ecosystem service typology (see, e.g., \cite{arias2018}, \cite{bela2015preliminary} and \cite{sagoff2008economic}). 

The claim made by VES that multiple values occur \say{naturally} within each ESF category is then consequential upon the fetishistic use of the value category immanent in the valorisation of value (see, e.g., \cite{degroot2002}, \cite{fisher2010valuing} and \cite{bela2015preliminary}). In such immanence, the valuation of varied values seems straightforward --- an action that requires no (thorough) examination. The price-tagging aspiration of monetary techniques is seen as an effort to signal the importance of nature in a way intelligible to the making of decisions, while the antinomic character of non-monetary values is expelled through the recognition of a multiplicity of values, in order to be later recuperated into the same decision-making rhetoric \citep{kosoy2010payments}. 

This unreflective usage furthers the fetishist forms of social praxis built around value, normalising VES as an exhaustive and objective description of our relation with nature \footnote{See the definition of total economic value in, e.g., \cite{pearce1994economic} and \cite{adger1995total}.}. Here a type of instrumental reason is at work, where, since the purposes of VES are taken for granted and self-explanatory, valuation practitioners need only consider the adequacy of their methods to said purposes. As these ends are not submitted to (categorical) reflection, and reflective thought is believed to come to its own when it rejects any status as absolute, reason devolves into a tool for coming to terms with the preestablished social praxis \citep{agamben1999potentialities}. 

Through this partial self-supression, instrumental reason, via VES, realises itself as the capitalist transformation of the world in its allegedly ecological facet. And in this process, instrumental reason becomes objective reason, for it reproduces the validity of the fetishistic treatment of value \citep{noys2010persistence}. Continued reflection at the categorical level --- such as ours --- is contested as speculative lucubration, precisely because the fetishistic relation has already been constituted as a natural necessity.  

\subsection{Inconsistencies, Contradictions and Value}

The validity of VES is judged on the basis of its conformity to the sociality from which it stems, implying, in turn, that the validity of its results is judged on their consistency with respect to the capitalist social praxis. It is as though objective reason would give way to an equally objective form of instrumental rationality \citep{jappe2017writing, hemmens2019critique}. In the context of VES --- and, most likely, in other domains of theoretical and practical activity \citep{kurz2019substance} --- \textit{(objective) reason is but the partial reflection on value and its dynamics}, whence \textit{rationality surges as a gauge of consistency with respect to said reflective/interpretive moment}. 

Rationality, in the previous sense, is expected from valuation practitioners as much as it is required from the subjects of valuation \citep{bela2015preliminary,barbier2009valuation,gomez2016concepts, jacobs2016new,kumar2008valuation}. Whether rationality substantiates in its laxer --- e.g., procedural rationality \citep{simon1976substantive,wakker2010prospect} and ecological rationality \citep{glymour2001mind, hogarth2007heuristic} --- or more stringent --- e.g., economic rationality \citep{simon1996economic} --- forms, it is considered the defining quality of (human) agency. Rationality is then posed as the conformity of means and ends prescribed by the process of the valorisation of value, thus subjectifying the abstract capacity to act in accordance to the reproduction of value --- thus (re)inaugurating the automatic subjectivity of value \citep{jappe2013sohn, hemmens2019critique}.

The assumption that rationality holds is equivalent to the hypothesis that the world is consistent \citep{routley1976dialectical} --- that our (capitalist) sociality is consistent. Yet more formally, the consistency hypothesis is the thesis that the class of all truths $T$ contains no contradictory pairs of the form $A$ and $\neg A$, where $\neg A$ is the negation of some predicate $A$ \citep{routley1976dialectical}. Of course, $T$ does not exhaust the class of all statements, for it is an empirical truth that the world is not trivial. Likewise, it is empirically true that the world is fraught with inconsistencies. We are, therefore, inclined to believe that inconsistency does not lead to triviality, and that its analytical appropriation, as opposed to its treatment, is better suited to achieve the break --- albeit partial --- from the internal contradictions of the social relations fostered by VES. 

The core inconsistency of our modern forms of socialisation lies in the fact that we have created economic laws, technological imperatives and similar structures revolving around value, which ended up shaping the conditions of the world, in a way that their mediating role has become our social actuality \citep{lukacs2017reification}. Put differently, value can only be produced by labour, yet labour is continually eroded by the reproduction of value, mainly by its estranging of labourers from their own creations (commodities) \citep{kurz2019substance}. Therefrom arises a more palpable inconsistency: the reduction of value contained in each commodity, induced by the erosion of labour, has to be compensated by an increase in the total mass of commodities produced, bringing about ecological consequences which will ultimately hamper said productive expansion.

VES practitioners ostensibly set out to treat the latter contradiction \citep{gomez2016concepts,gomez2015ecological,bela2015preliminary,badura2016valuing}. They intend to elicit value from nature, where they believe lays dormant, so as to better inform the commodity production system \citep{adger1995total,arias2018}. VES is, from their perspective, a mechanism to do away with the contradiction.  

Not persuaded by the belief that all that exists is in some way valuable, for we recognise that the underlying rationality of this credence is value-based by construction, we adopt a negative dialectical approach to valuation. Instead of treating inconsistencies, we operate on them, not necessarily to transcend them, but to discern other forms of sociality \citep{adorno1973negative}. This determination will not only have to deal with the fact that value relations are inconsistent, but also acknowledge that the inconsistency of value relations is not trivial, which, in turn, implicates that the classical logic underpinning instrumental rationality is inadequate to fulfill these requirements \citep{priest2008introduction}.

Value relations being inconsistent means that they engender contradictions, and that these contradictions are (empirically) valid. Value relations being non-trivial (while inconsistent) means that not everything follows from the recognition of their internal contradictions. Classical logic, as the cornerstone of rationality, does not allow for contradictions, or whenever present they provoke the explosion of the analytical model sitting on top, in the sense that, by effect of the disjunctive syllogism, the validity of any other premise can be deduced from an inconsistent argument \citep{priest2008introduction, berto2007sell}. For instance, it is perfectly classically rational to accelerate economic growth if it is the case that the production of commodities increases, and the biophysical conditions underpinning such increment deteriorate; however, the coexistence of both premises is not valid within the same classical framework.  

Our critique embraces contradictions and inconsistency, but it is not irrational. It is not irrational in the sense that Aristotle's principle of non-contradiction is (should be) correct in our foregoing proposal, both in its syntactical and semantical formulations \citep{wedin2004use}. That is, using the $\land$ symbol to denote conjunction, Aristotle's principle $\neg (A\land \neg A)$ is valid, hence true, as is the statement indicating that no statement is both true and false, which, as will be shown in Section \ref{sec:logics_contradiction}, is guaranteed by the bivalent features of our model's semantics. Thus, we join Routley in claiming that there are genuine contradictions in things \citep{routley1976dialectical}. 

In Section \ref{sec:logics_contradiction} we will construct a methodological framework conducive to a logical model that allows for inconsistency and contradictions --- a paraconsistent and dialetheic one. The resulting model will be at the centre of a computational environment for the reproduction of the dialogues, determining how the Páramo is appropriated based on the notion of value presented hitherto. Before detailing the methodological upshot of our considerations, a brief account of the Guantiva-La Rusia Páramo is in place.

\section{The Guantiva-La Rusia Páramo}
\label{sec:paramo}
The Andes contain some of the most abrupt environmental contrasts. They rise from the sea coast to volcanoes and glaciated massifs, and then descend rapidly into the tropical Amazon Basin \citep{brush1982natural}. In Colombia, these transitions occur on a horizontal distance of 400 km \citep{parra2009episodic}. The great volume of compressed material within such a narrow space has produced varied landscapes of great diversity: intermontane valley floors covered in some places with tropical forests and in others with desert vegetation, which ascend steeply, through temperate forest belts, to páramo or permanent snow \citep{cortes2013vision}.

The páramo is an alpine biome characteristic of the high tropical Andes, somewhat arbitrarily, defined as those grassland ecosystems occurring above 3000 MASL \citep{rivera2011guia}. Geologically, these areas are relatively young and have been subjected to the effects of glaciations which resulted in cycles of morphological fragmentation and reconnection \citep{flantua2019flickering}. This flickering connectivity explains the unique flora of the páramo and its pivotal role in the cycling of water \citep{sugden2019paramo}. 

The vegetation structure of the páramo is one of the most complex among grasslands \citep{hofstede1995effects, ramsay2014giant}. The dominant growth forms of the páramos, along the Colombian Andes, are stem rosettes (Espeletia spp.), constituting an emergent vegetation layer above high (up to 1 m) growing tussock grasses and dwarf shrubs \citep{diazgranados2017espeletia}. An additional lower vegetation layer may be present with ground rosettes, sedges and short-growing grass species \citep{diazgranados2017espeletia}.

Stem rosettes are plants with the caulescent rosette growth form \citep{smith1979function}. These plants have unbranched or semiwoody stems supporting rosettes of evergreen leaves, which are generally marcescent, remaining on the stem after they senesce \citep{smith1979function}. It is probable that the caulescent rosette form provides a buffering effect against the diurnal fluctuations in temperature and the seasonal fluctuations in rainfall, allowing the plant to trap moisture through its dense pubescence \citep{diazgranados2017espeletia}.

It has been observed that dew accumulation in the rosettes is sufficient to cause stem flow, increasing soil moisture. Sifting of fog and rain by the rosette leaves would have a greater effect \citep{hofstede1995effects}. If the Espeletia's leaves absorb enough moisture, a net movement of water from leaves to roots, and then to soil, might occur, provided the soil water potential is significantly lower than the plant water potential \citep{hofstede1995biomass}.

Although the exact mechanisms and processes through which the páramo vegetation contributes to the regulation of water are not fully understood, the low levels of evotranspiration induced by the vegetation's capacity to intercept and store water are judged crucial in the cycling of water toward adjacent river basins \citep{buytaert2006human}. It is the much better known structure of the páramo soils and accidented toporgaphy, of rough steep valleys and almost flat plains, the factors commonly invoked to explain the páramo's high water regulating capacity \citep{buytaert2006human}. In conjunction, these properties are believed to explain an average net contribution of water to local hydrological systems of approximately 1400 mm, or 66.5 $km^3$, per year \citep{hincapie2002transformacion}. 

\subsection{The Guantiva-La Rusia Paramunos}

 \begin{figure}
	\centering
	\begin{minipage}{0.5\textwidth}
		\centering	
		\includegraphics[width=\textwidth]{example-image-a}\\ 
    	\caption{GLR Complex Map}
    	\label{fig:map}
	\end{minipage}
	\hspace{-0.5cm}
	\begin{minipage}{0.5\textwidth}
	\centering
    \resizebox{6cm}{4cm}{
    	\begin{tabular}{M{2cm}|M{2.5cm}|M{2cm}|M{2cm}}
    		\toprule
    		\textbf{Province} &  \textbf{Municipality} & \% Area within the Complex & \% Area within the Municipality\\
    		\midrule
    		\multirow{7}{*}{\textbf{Santander}} & Encino  &   37 &   12 \\
    		& Onzaga   &   30 &   12\\
    		& Coromoro   &   23 &    11\\
    		& Charalá   &   3 &   0.9\\
    		& San Joaquín &  2  &  0.3\\
    		& Gámbita   & 0.2 &  0.1\\
    		& Mogotes   &  0.1 &  0.02\\
    		\hline
    		\multirow{16}{*}{\textbf{Boyacá}} & Tutazá  &  76 &   8 \\
    		& Belén   &   54 &   7\\
    		& Sativanorte   &   53 &    7\\
    		& Duitama   &   52 &   10\\
    		& Susacón  &  51 &   8\\
    		& Cerinza &  49 &   3\\
    		& Sta. Rosa de Viterbo &  43 &  4\\    		
    		& Sativasur &  32 &  1\\
    		& Paz de Río &  205 &  155\\
    		& Soata &  23 &  2\\
    		& Beteitiva &  22 &  2\\
    		& Sotaquirá &  12 &  3\\
    		& Floresta &  12 &  0.9\\
    		& Paipa &  9 &  2\\
    		& Nobsa &  8 &  0.4\\
    		& Tipacoque &  6 &  0.4\\
    		\bottomrule
    	\end{tabular}
    	}
    	\caption{Area Distribution of the Guantiva-La Rusia}
    	\label{tab:percent_within}  
	\end{minipage}
	\end{figure}

On the western side of the eastern branch of the Colombian Andes raises the Guantiva-La Rusia páramo complex, spreading over several municipalities of the provinces of Santander and Boyacá. Judging by its more evident biophysical features, the altitudinal range of the complex varies from 3100 MASL up to 4280 MASL \citep{cortes2013vision}. However, the recognition of its role in the provisioning of water is prevalent among the people living within its buffer zone, and even in areas well below this belt of cloud forest \citep{rivera2011guia} --- communities which, along with those inhabiting the páramo, we term \textit{Paramunos}. 

\enlargethispage{1\baselineskip}

The locations circumscribing the Guantiva-La Rusia include the municipalities of Charalá and Encino in Santander, and also the municipalities of Duitama and Belén in Boyacá. Table \ref{tab:percent_within} presents the distribution of the area of the páramo complex among the different municipalities, as per the official delimitation criteria \citep{rivera2011guia}. Duitama, Paipa, Belén, Charalá and Encino, despite not having the largest share of the total páramo area, concentrate the economic activity of the region \citep{rojas2017caracterizacion}. 

Until not so long ago, dairy farming, cattle raising, wood extraction --- from endemic oak (\textit{Quercus Humboldtii} spp.) forests --- for charcoal production, and the cultivation of potatoes constituted the bulk of the economic activities within the complex \citep{rojas2017caracterizacion}. These activities typically involved the clearing of large areas of páramo and forest, through burning or grazing at regular intervals of one to several years \citep{rojas2017caracterizacion}. More recently, the demographic changes induced by the mediation of value relations, in the sense of Section \ref{sec:rationality}, reduced the intensity of these practices, while the appropriation of the relations themselves produced a move away from them.

Burning, grazing, and cultivating above the 3000 MASL threshold are now considered an affront to collective interests, for they have been observed to cause negative impacts on the water regulating capacity of the páramo, via structural changes in its soils \citep{buytaert2006human}. So, not only are these activities little effective on páramo lands, inherently devoid of a farming vocation, but also detrimental to the regional water yield. Traditional practices are proscribed to a handful of campesinos forced to live off the higher lands; nonetheless, a scenario where these practices become once again widespread is always likely. 

Some of the reasons behind the ostensible imminence of this regression reside in the land-holding patterns of the Colombian rurality. In particular, the competition between large states (\textit{haciendas}) and peasant communities --- a common and often violent feature of rural Colombian society --- can be recognised in the way Paramunos lead their lives \citep{guzman2019violencia}. Although bellicose manifestations of violence are not apparent in the Guantiva-La Rusia complex, institutionalised forms of violence have had a similar effect. The result being the historical concentration of land ownership, the uneven distribution of population densities and land-use pressures, and the absorption of displaced landless labourers and peasants into semi-feudal or proto-capitalist systems of agricultural production \citep{guzman2019violencia}. 

This legacy of inequality and the more recent changes in the páramo sociality are now being incorporated into a rationale of the sustainable use of nature and the preservation of traditional forms of living, typically associated with the \textit{campesino} identity \citep{rivera2011guia}. Previous farming practices are eschewed in favour of modes of production and livelihoods compatible with the rhetoric of conservation. Haciendas are now seen as private conservation areas, schemes of payments for ecosystem services are slowly becoming part of the locals' imagery as a means to avoid resuming outdated practices, while collective initiatives are being touted as profit-making enterprises in niche markets for \say{environmentally friendly} or \say{ethical} commodities. 

The communities of the páramo, the cloud forest, and adjacent lower areas --- which altogether we denote \textit{Páramo} --- are undergoing two major changes. On the one hand they are committed to transitioning from traditional agropastoral economies towards entreprenurial activities centered around agro/eco-tourism. And on the other, they strive to reinterpret (preserve) campesinos as local subjects of the (world) market. 

These tendencies are not being forcefully imposed upon the people of the Páramo \citep{rojas2017caracterizacion}. They constitute a conscious determination on how the social relations of the Páramo are to be shaped. Notwithstanding the repercussions of neoliberal policies in Colombia, it is the narrative of developmental sustainability that serves as the backdrop for their enactment \citep{rivera2011guia}. Markets, for campesino products and services, are seen as a medium for channeling revenues while incentivising the conservation of nature and the preservation of a way of life. These notions, however, are not subject to critical consideration. 

During our time with the Paramunos, we noticed that the invocation of markets is grounded on the belief that value --- allegedly inherent in all (their) activities; productive or otherwise --- can be adequately represented in a monetary form. Several concerns were voiced regarding the necessity to monetise traditional knowledge, and receiving a fair compensation for the conscious conservation of the Páramo. We recognised a view that essentialises value and sees in its valorisation, through the participation in (new world) markets, a medium for the fulfilment of material needs and aspirations.

\subsection{Value in the Páramo}

The situation of the Paramunos is an instance of the reproduction of the capitalist social praxis presented in Section \ref{sec:rationality}. Value is considered an existential --- not even anthropological --- constant, a fundamental property of all that exists which can be understood without reference to capitalism and its historical specificity as a mode of life. This ontologisation of value enables value fetishism and its aporetic effects, manifested as the subjetification of the Paramunos in terms of the more global campesino identity, the re-signification of nature as a portfolio of goods and services, and the re-coding of the collective as fundamentally entrepreneurial. 

Value, in this sense, becomes a transhistorical invariant waiting to be realised through market mechanisms, i.e., prices \citep{kurz2019substance}. Not much thought goes into the nuances of the implicit recognition that value is, in principle, intrinsic, and how labour can augment/create it, or how value's elicitation may have a tinge of subjective consideration \citep{jappe2017writing}. Regardless, nature harbours value; and that is simply a self-evident fact. However imperfect, prices are but neutral means of expressing nature's value, of expressing the value that Paramunos --- re-imagined as campesinos --- create, and of unifying the overwhelming subjectivity of these various values. 

The Paramunos' approach to value coincides with VES' prescription on value elicitation (see, e.g, \cite{gomez2016concepts}, \cite{gomez2015ecological},\cite{bela2015preliminary}, and \cite{badura2016valuing}). The ESF typology would, then, seem apropos when discussed in the context of the Páramo. The growing of coffee, fruits, Andean legumes and tubers embodies provisioning services; the stewardship of Espeletia and oak forests incarnate (hydrological and carbon sequestration) regulating services; and agro-touristic activities, alongside the efforts to preserve the campesino identity, typify cultural services. Every form of appropriating the Páramo constitutes a commodity whose nascent market requires but the assistance of VES practitioners to adequately estimate their corresponding prices. Once set, prices will harness the Paramunos' predisposition to rational behaviour, eventually attaining efficiency in the use of nature, for the allocation of property rights is, more or less, irrelevant to market efficiency \citep{coase1960problem, farley2010payments}.

The Paramunos' reasoning abides by the type of instrumental rationality described in Section \ref{sec:rationality}. Their reflective considerations on value are rational, because they coincide with the eminently transactional ideal of VES, where individual preferences are equalised through pricing in a self-fulfilling vision of unity. Rationality is, consequently, established as the confirmation of a preordained social praxis that encourages circulation over production.

Unlike our labour-based definition, VES' notion of value is rooted in the sublimation of the sphere of circulation of commodities. Accordingly, value is posited a fundamental property of all things, liable to being reproduced and augmented, but realisable only through exchange and circulation --- value is subjective; that is, solely dependent on individual preferences. Whence multiple values arise, and an arbitrary social convention of arbitrage, i.e., prices, surge as an equaliser of this impractical overabundance. But for prices to substantiate, circulation must be countenanced. Therefore, it makes complete sense to promote the creation of markets even for unlikely commodities which, in principle, cannot be produced, as in the hydrological regulating services \say{provided} by the páramo vegetation.

In contrast, a labour-based explication of value treats it as an objective property born of the material appropriation and transformation of the world, justified by the experiential primacy of production over circulation \citep{kurz2019substance, hemmens2019critique, marx2019capital}. Labour-based value is also manifested through exchange, but with reference to the amount of time expended in the manufacture of reality, i.e., abstract labour, that a commodity contains \citep{marx2019capital}. Here, pricing does not occur as arbitrarily, and the reproduction of value (the valorisation of value) takes a hold of its material substratum. Value fetishism also becomes more readily apparent, for the recognition of the material and historical basis of value delineates the ambit where exchange takes place. 

This labour-based exegesis makes clear that VES' ontologising approach to value uses ESF to extrapolate value fetishism to the Paramunos' relation with nature, by way of extending the ambit of exchange and of the valorisation of value. In analysing this process we reiterate, first, that Paramunos do not conform to this approach by diktat nor by effect of false consciousness, and, second, that said conformity is not objectionable in itself but rather undesirable based on its detrimental effects on the Páramo. The former observation can also be read as evidence of rational behaviour on the part of the Paramunos, while the latter refers to the practical consequences of reproducing the contradictions of the capitalist social praxis transmitted through the furthering of value fetishism. More to the point, we are referring to the advancement of alienation implicated by the adoption of the campesino identity at the expense of the paramuno lore, and other unintended environmental impacts that the tertiarisation of already precarious economies may have on the páramo.

\enlargethispage{2\baselineskip}
In the next section, we use an example to illustrate the relevance of a new formalism --- presented in Section \ref{sec:logics_contradiction} --- for addressing the issues raised here. This example is based on a classical logical framework which allows for instrumental rationality in the use of the páramo, as per the ESF. The shortcomings of this approach --- e.g., the impossibility of handling its inherent contradictions --- uphold the need for a more comprehensive logical framework, and a distinct perspective on conducting VES, which, we demonstrate, is enabled by our proposal.  
\section{The Logic of the VES Rationality}
\label{sec:example}

\begin{figure}
	\centering
	\includegraphics[width=0.5\linewidth]{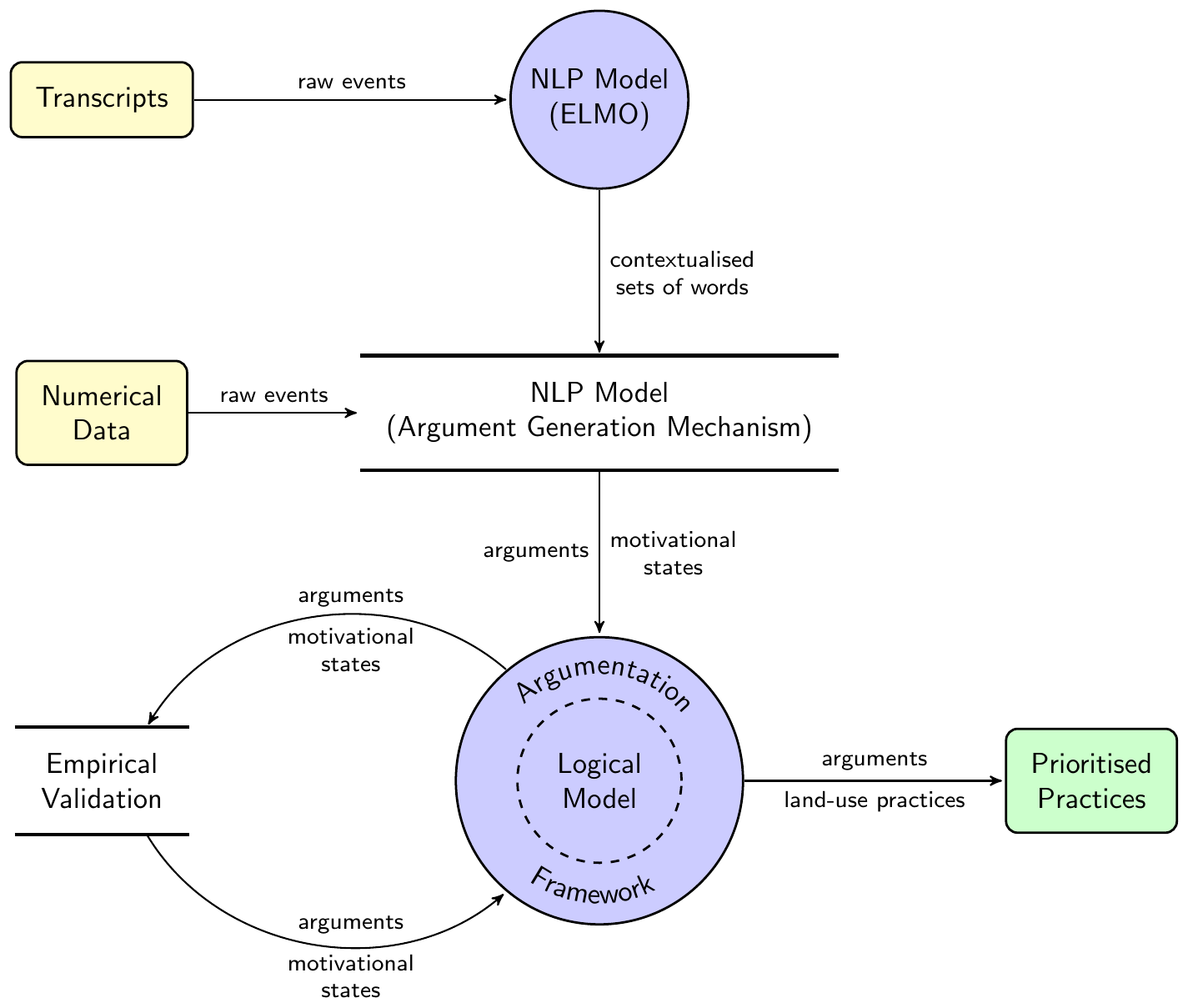}
	\caption{High-level Description of our Computational Environment}
	\label{fig:av}
\end{figure}

Before looking at our example, let us give a brief indication of what we mean by computational environment. By computational environment we understand a collection of computational techniques that (re)produce dialogical settings where multiple actions on the Páramo are, intentionally or otherwise, conflated with the notion of value. These techniques delineate an environ for all interested stakeholders to put forward their views on how the Páramo can be construed, and pit these arguments against one another as a means to prioritise over multiple, and often conflicting, management practices. 

The core of our approach resides in its paraconsistent logical model, in the sense of Section \ref{sec:rationality}, wrapped up in an argumentation framework. Argumentation frameworks are abstract devices that formalise dialogues \citep{Walton2009}. They are composed by a body of arguments, a relation of attack stating which arguments can pose a challenge to some other arguments, and a set of motivational states underpinning the framework's arguments \citep{prakken2018historical}. The argumentation framework we use is special for at least two reasons: 1) it incorporates information on which and under what conditions the outcome of an argument is preferred over distinct alternatives, and 2) it determines the validity of a given argument through a logical model of entailment. Definition \ref{def:vaf} provides a formal account of a value-based argumentation framework (VAF).

\begin{definition}[\textbf{Value-based Argumentation Framework} \citep{bench2009abstract}\label{def:vaf}] A value-based abstract argumentation framework is a tuple $\mathcal{VAF}=\langle \mathcal{A},\mathbf{R},V,val,P\rangle$. $\mathcal{A}$ is a set of arguments, $R\subseteq \mathcal{A}\times\mathcal{A}$ is a binary relation of \textit{attack}, $V$ is a non-empty set of \textit{values}, $P$ is the set of total orders on $V$, and $val:\mathcal{A}\mapsto V$. 
\end{definition}

The logical model of our approach establishes the low-level conditions to infer whether an argument is defensible within a particular dialogical setting \citep{besnard2009argumentation}. A task that is accomplished while allowing for inconsistent arguments, i.e., arguments involving both a premise and its negation, but do not entail trivial outcomes, as outilned in Section \ref{sec:rationality}. It is in this sense that our logical model is said to be paraconsistent \citep{berto2007sell}.

The information on which the logical model and the argumentation framework operate is generated by a natural language processing (NLP) model. The main function of the NLP model is to facilitate the use of raw text data in the form of natural language and numerical data alike. Our NLP model is made up of two parts: an \textit{embeddings from language model} (ELMO) which learns a (numerical) vector representation of each word per the word's sense/context \citep{peters2018}, and an argument generation mechanism that arranges these word embeddings into a format amenable to VAF. Although, from a conceptual perspective these two processes may be considered separately, from a functional point of view they are integrated into a single NLP model \citep{manning1999foundations}.

The previous description is more succinctly illustrated in Figure \ref{fig:av}. It shows how the NLP model synthesises unfiltered raw data ---the transcripts from interviews and other synthetic or observed events of numerical nature--- into proto-arguments and their underlying motivational states, which are then fed to the argumentation framework. The argumentation framework and the logical model's proof theory would then generate a list of defensible arguments which will be contrasted in the field, and, if not discarded, would then returned enriched to the argumentation framework, and an order of prioritised practice is generated. 

The following example presents our computational environment at work. It illustrates the functioning of a VAF, and the need for a paraconsistent model to reason about interactions revolving around the use of the Páramo and related value-based categories.

\subsection{Classical Logics and ESF}

Our computational environment is based on the logical formalisation of dialogical interactions. As originally envisioned, its logical formalism should be non-classical. That is, our logical model should contain non-trivial theses of the form $\{a \land \neg a\}$, where $a$ is a premise belonging to the support of a generic argument $A\in\mathcal{A}$, and $\mathcal{A}$ is as in Definition \ref{def:vaf}. 

Current research into logics amenable to structural contradictions --- within the computational argumentation domain --- is still incipient \citep{amgoud2010formal, arioua2017logic, prakken2018historical}. For this reason, this section is limited to elucidating what is meant by model of logical entailment in the context of our proposal, and how it is articulated into the a VAF by way of an example. The justification of a non-classical model of entailment is given in Section \ref{sec:rationality}, whereas its integration into a working (counter-)valuation framework will be investigated in Section \ref{sec:logics_contradiction}. Let us, then, introduce a classical logic-based instance of our computational environment through Example \ref{ex:main}.

\begin{example}[]\label{ex:main}
Our example represents the predicament between using páramo sites for agricultural purposes and performing restoration activities involving the same areas. Our variables of interest are designated as follows:

\begin{description}
\item[$a:$] Increased agricultural land-use.
\item[$h:$] Improved hydrological regulating services.
\item[$r:$] Greater number of peatland restoration activities.
\item[$s:$] Increased water supply.
\item[$y:$] Greater agricultural yield.
\item[$w:$] Improved living conditions.
\end{description}

To construct our logic-based VAF we make use of sentential logic with a conventional linguistic structure. The resulting construct is considered a classical logic \citep{buning1999propositional}. This classical logic consists of well-formed formulae built up from sentential variables --- such as $h$ or $y$ --- \if and sentential constants --- such as punctual levels of $h$ or $y$ ---\fi using the sentential connectives $\to$ (implies), $\land$ (and), $\lor$ (or), and $\neg$ (not). Every argument in our VAF is formed by combining these sentential elements.

We start off by defining a knowledge base, which summarises all relevant information regarding our variables:

$$KB_0 := \{a, a\to \neg h,a\to y, h\to w, r, r\to\neg a, r\to\neg y, r\to h, y\}$$

The main or \textit{root} argument will be that of prioritising agricultural land-use, as it leads to a greater yield (and higher income), i.e.,

$$A_1 := \langle \{a, a\to y\}, y\rangle.$$

In contrast, restoration activities could be directly advocated --- agricultural land-use is eschewed in favour of peatland restoration activities to improve water quality, i.e.,

$$A_2 := \langle\{r, r\to \neg a\}, \neg(a\land (a\to y))\rangle;$$

which, in turn, could be refuted by the primacy of the income from agricultural activities over non-monetary considerations, i.e.,

$$A_3 := \langle\{y, y\to \neg r\}, \neg (r\land (r\to \neg y))\rangle.$$

Yet, another argument may affirm that restoration is desirable as it guarantees the regulating services hampered by agricultural activities, i.e.

$$A_4 := \langle\{(r\to h)\to (h\to \neg a)\}, \neg(a\land (a\to y))\rangle.$$

In consequence, the abstract argumentation framework of our example takes the form

$$\mathcal{AF}_{KB_0}=\{\{A_1, A_2, A_3, A_4\}, \{(A_1, A_2), (A_1, A_4), ( A_2, A_3)\}\};$$

and its corresponding argumentation tree appears as in Figure \ref{fig:arg_tree_1}.

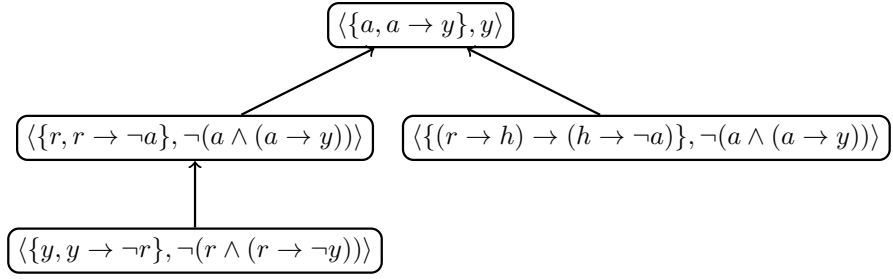
\begin{figure}[h]
    \centering
    \begin{tikzpicture}[sibling distance=6cm, every node/.style = {shape=rectangle, rounded corners, draw, align=center}, line width=.3mm]]
    \node {$\langle \{a, a\to y\}, y\rangle$}
        child { node {$\langle\{r, r\to \neg a\}, \neg(a\land (a\to y))\rangle$} edge from parent[<-]
            child { node {$\langle\{y, y\to \neg r\}, \neg (r\land (r\to \neg y))\rangle$} edge from parent[<-]}}
        child { node {$\langle\{(r\to h)\to (h\to \neg a)\}, \neg(a\land (a\to y))\rangle$} edge from parent[<-]};
    \end{tikzpicture}    
    \caption{Argumentation Tree for $\mathcal{AF}_{KB_0}$}
    \label{fig:arg_tree_1}
\end{figure}{}

To extend our current AF to a value-based framework we recognise $V=\{y, w\}$ as our set of values, in view that $a\to y$ and $(r\to h)\to (h\to w)$. So, there are two possible audiences $P=\{y\succ w, w\succ y\}$ with respect to the binary relation $\succ$ on $V$, which we designate as audience-$y$ associated with order $y\succ w$ and audience-$w$ associated with order $w\succ y$. Their respective variables are tagged with the subscript $(\cdot)_i$ for $i\in\{y,w\}$, whenever the latter's omission may seem ambiguous. Finally, note that $val=\{A_1\mapsto y, A_2\mapsto w, A_3\mapsto y, A_4\mapsto w\}$ and our VAF can be expressed as follows\footnote{The symbol ``$\mapsto$" is used  to represent the mapping from arguments to values, as per Definition \ref{def:vaf}. It is distinct from the implication ``$\to$".}

{\small\begin{align}
	\begin{split}
		\mathcal{VAF}_{KB_0}&=\{\{A_1, A_2, A_3, A_4\}, \{(A_1, A_2), (A_1, A_4), (A_2, A_3)\}, \{y, w\},\\
		& \{A_1\to y, A_2\to w, A_3\to y, A_4\to w\}, \{y\succ w, w\succ y\}\}
	\end{split}
	\end{align}}
    
The Hasse diagrams in Figure \ref{fig:hasse_1} offer an alternative representation of the preferred extensions of $\mathcal{VAF}_{KB_0}$\footnote{see \cite{afanador2019arguing} for a definition of preferred extensions as a solution concept of VAFs.}. It indicates that the preferences contained in our VAF are uninformative, as we end up with two independent sets of preferred arguments for each audience. Put another way, the values associated with these arguments are trivial: audience-$y$ favours agricultural activities over restoration as opposed to audience-$w$, precluding finding a common ground. However, if the knowledge base is extended to include premises stating that ``increased water supply generates greater agricultural yields", and that ``improved regulating services uphold an increased water supply", i.e., $KB_1\equiv KB_0\cup\{s\to y, h\to s\}$, then a new preferred extension arises.

\begin{figure}
    \centering
    \begin{tikzpicture}[sibling distance=6cm, line width=.3mm]]
    \node [circle, draw] (o) {$a$}
        child {node [circle, draw] {$r$} edge from parent[->]};
    \node[above of= o,fill=white,inner sep=.5pt, line width=2pt] {audience-$y$:};
    \end{tikzpicture}
    \hspace{2cm}
    \begin{tikzpicture}[sibling distance=6cm, line width=.3mm]]
    \node [circle, draw] (o) {$r$}
        child {node [circle, draw] {$a$} edge from parent[->]};
    \node[above of= o,fill=white,inner sep=.5pt, line width=2pt] {audience-$w$:};
    \end{tikzpicture}
    \caption{Hasse Diagram for $\mathcal{VAF}_{KB_0}$}
    \label{fig:hasse_1}
\end{figure}
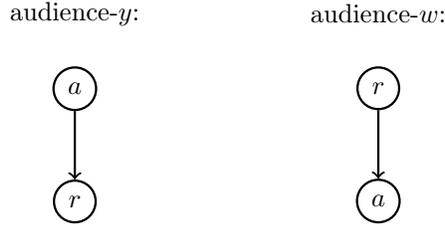{}

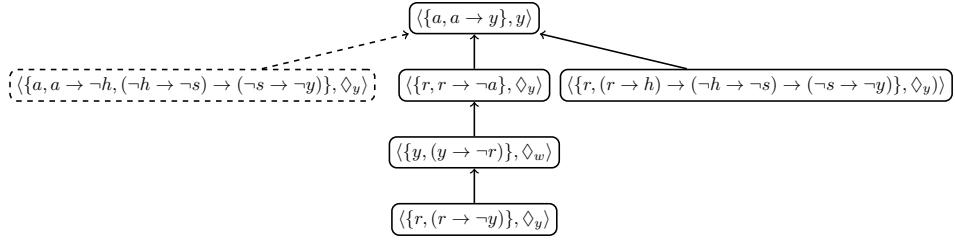
\begin{figure}
	\centering
 	\scalebox{.7}{
 	\begin{tikzpicture}[sibling distance=6.3cm, every node/.style = {shape=rectangle, rounded corners, draw, align=center}, line width=.3mm]
 	\begin{scope}[scale=.85]
 	\node {$\langle \{a, a\to y\}, y\rangle$}
 	child { node[dashed] {$\langle\{a, a\to \neg h, (\neg h\to\neg s)\to (\neg s\to\neg y)\}, \lozenge_y\rangle$} edge from parent[<-, dashed]}
 	child { node {$\langle\{r, r\to \neg a\}, \lozenge_y\rangle$} edge from parent[<-]
 		child { node {$\langle\{y, (y\to \neg r) \}, \lozenge_w\rangle$} edge from parent[<-]
 			child { node {$\langle\{r, (r\to\neg y)\}, \lozenge_y\rangle$} edge from parent[<-]}}}
 	child { node {$\langle\{r, (r\to h)\to (\neg h\to\neg s)\to (\neg s\to\neg y)\}, \lozenge_y)\rangle$} edge from parent[<-]};
 	\end{scope}
 	\end{tikzpicture}}
  	\caption{Argumentation Tree for $\mathcal{VAF}_{KB_1}$}
	 \label{fig:arg_tree_2}
\end{figure}

The changes $KB_1$ introduces are reflected in a new argumentation tree; that of Figure \ref{fig:arg_tree_2}. A lighter notation has been adopted in Figure \ref{fig:arg_tree_2}, by making $\lozenge_y:=\neg(a\land(a\to y)$ and $\lozenge_w:=\neg(r\land(r\to\neg a)$. For the sake of brevity, instead of deriving every element in $\mathcal{VAF}_{KB_1}$, let us simply note that, while the sets of values, and audiences from $\mathcal{VAF}_{KB_1}$ are preserved, there are two new counter-arguments to the root, altering the value mappings between $V$ and $P$. One affirming that the dedication of land to agricultural activities negatively impacts hydrological regulating services, reducing the water supply, the agricultural yield, and eventually eroding the livelihoods of the locals, i.e., 

$$A_5:=\langle\{r, (r\to h)\to (\neg h\to\neg s)\to (\neg s\to\neg y)\}, \lozenge_y\rangle,$$

and another stating that by impacting the water supply through regulating services, agricultural activities could eventually reduce the agricultural yields, i.e.,

$$A_6:=\langle\{a, a\to \neg h, (\neg h\to\neg s)\to (\neg s\to\neg y)\}, \lozenge_y\rangle,$$

an argument which will not be factored into the derivation of the preferred extension of $\mathcal{VAF}_{KB_1}$ --- hence, the dashed contours --- but one that will serve to motivate our discussion about the non-classical counterpart of the approach embodied in this example.

\begin{figure}[h]
    \centering
    \begin{tikzpicture}[sibling distance=6cm, line width=.3mm]]
    \node[circle, draw, label={\small $a\to y$}] (o) at (0,0) {$a$};
    \node[circle, draw, label={\small $(h\to s)\to(s\to y)$}] (p) at (2.5,0) {$r_y$};
    \node[circle, draw, label={below:\small $h\to w$}] (q) at (1.25,-2) {$r_w$};
    \node[fill=white,inner sep=.5pt, line width=2pt] at (1,1.5) {audience-$y$:};
    \draw[-] (o) to (q);
    \draw[-] (p) to (q);
    \end{tikzpicture}
    \hspace{2cm}
    \begin{tikzpicture}[sibling distance=6cm, line width=.3mm]]
    \node[circle, draw, label={\small $h\to w$}] (r) at (0,0) {$r_w$};
    \node[circle, draw, label={\small $(h\to s)\to(s\to y)$}] (s) at (2.5,0) {$r_y$};
    \node[circle, draw, label={below:\small $a\to y$}] (t) at (1.25,-2) {$a$};
    \node[fill=white,inner sep=.5pt, line width=2pt] at (1,1.5) {audience-$w$:};
    \draw[-] (r) to (t);
    \draw[-] (s) to (t);
    \end{tikzpicture}
    \caption{Hasse Diagram for $\mathcal{VAF}_{KB_1}$}
    \label{fig:hasse_2}
\end{figure}
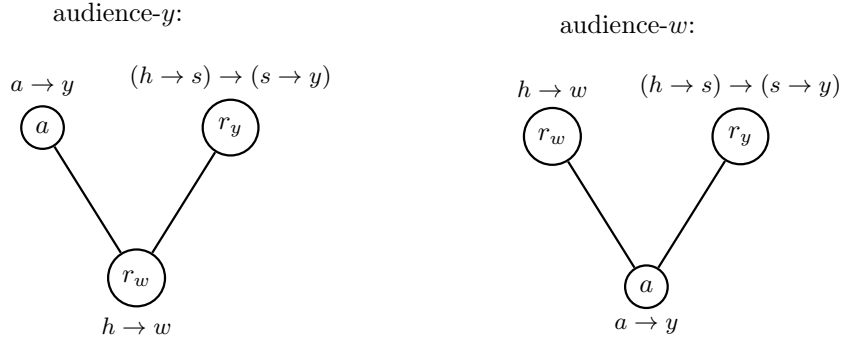{}

Like before, Hasse diagrams are used to represent preferred extensions. The Hasse diagrams of $\mathcal{VAF}_{KB_1}$ appear in Figure \ref{fig:hasse_2}. For audience-$y$, the preferred extension is $\{A_1, A_5\}$. For audience-$w$, the preferred extension is $\{A_2, A_5\}$. The observation that restoration has a beneficial, but indirect effect on agricultural yields, reconciles the two audiences through argument $A_5$. The final result being that restoration activities should be prioritised over agricultural ones.

This example comes to an end with a passing remark on argument $A_6$. Argument $A_6$ affirms that an insistence on promoting agricultural activities may be self-defeating, in view of the interdependency of the circumscribing ecological processes. Put another way, $A_6$ introduces the non-trivial thesis $IC:=\{a\land\neg a\}$. 

Given the notions of defeat, acceptability and connected ideas contained in the definition of a preferred extension (see \cite{afanador2019arguing}), it would not be possible to find an admissible argument for either audience-$y$ or audience-$w$ if theses like $IC$ were allowed into $\mathcal{VAF}_{KB_1}$. The reason behind this is that theses like $IC$ involve determining the truth value of contradictions, and the acceptance of paradoxes as proofs, tasks for which the numerous classical semantics are not equipped \citep{buning1999propositional}. Since inconsistency arises even in the commonplace observation that ecosystem services are interlinked, recasting VAFs to allow for non-trivial inconsistencies intends to explicate logically a non-negligible part of the complexity of VES associated with the furthering of value fetishism.
\end{example}

\subsection{The Need for a Paraconsistent Approach}
Example \ref{ex:main} shows the advantages of logically modelling the formation of arguments and their probing in the context of our computational environment. The example, however, does not present a complete application of our approach, in that it cannot cater for the inconsistencies arising from the internal contradictions of ESF and the value-based relations contained in the corresponding knowledge base. 

The tentative quality of our example is not a matter of incremental development, or a case of missing analytical pieces in the making. Our approach is incomplete, for its non-classical logical rooting \textit{is} its explanatory power. Even though VAFs furnish the environment where high-level representations of values and preferences take hold, the logical and concrete antinomies of conflicting forms of appropriating the Páramo --- induced by value-based relations --- can only be fully appreciated through a logical account of their inconsistency; or rather, their paraconsistency. Thus, Example \ref{ex:main} is an effort to thematise this conjecture, and by doing so it also illuminates some of the constitutive elements of our approach, for the set of defensible arguments it entails, and their connected actions, operate on and prefigure their underlying contradictions.

The next section introduces the paraconsistent, and dialetheic (see Section \ref{sec:rationality}), model of logical entailment at the heart of our approach. It substantiates our critical stance on VES and ESF. We examine a viable analytical extension to VAFs leading, rather, to a different class of mathematical objects --- semantics games --- and proceed to illustrate their capacity to deal with contradictions arising from ESF by way of two examples. The section comes to an end with a brief discussion on the rationality of our critique and its methodological manifestation.
\section{The Logics of Contradiction}
\label{sec:logics_contradiction}

In Section \ref{sec:rationality}, we showed that the ontologisation of value is sustained by the assumption that freedom of contradiction is a crucial determinant of reality, understood as the reproduction of the capitalist social praxis. However our practical experience suggests otherwise, thus impelling the abandonment of this thesis. This section aims to corroborate that such observation, and the critique of the value-based categories from which it stems, are not irrational. We do this by constructing a game-theoretic object upon a logical model of entailment endowed with a dialectical fact, which makes up for the shortcomings of the classical model in Example \ref{ex:main}. 

Two main ideas were drawn from our discussion in Section \ref{sec:rationality} and our analysis of Example \ref{ex:main}. First, we concluded that freedom from contradiction does not provide an acceptable necessary condition for rationality, rational belief, or rational inquiry. Second, we established that, by allowing for an analytical framework amenable to inconsistencies, there might be a way out of the fetishist constitution of value in VES; or, equivalently, that by advancing the main points of our critical approach to valuation, a viable form of methodological counter-praxis may arise.

\subsection{Paraconsisent Dialogues on Value as Dialectical Games}

What we want to do now is to incorporate the above conclusions \iffalse on how contradictions can be semantically treated\fi into our computational environment, in order to conduct a (paraconsistent) discursive analysis in line with our observations in Section \ref{sec:example}. We accomplish this by analysing dialogues in the form of an strategic interaction infused by the notion logical significance embedded in a \textit{d-model} of entailment --- a logical description of logical relations based on meaning, as presented in \cite{anderson2017entailment}. We use a game-theoretic formalism to tie up all these ideas together. 

\begin{definition}[\textbf{Dialectical Dialogical Game (DDG)}]
    A dialectical dialogical game is a tuple $\G=\langle \D, N; \rho, \sigma\rangle$, where $\D$ is a d-model, $N$ indicates the number of players, $\rho = \{\rho_G, \rho_L\}$ is a well-defined set of rules, where $\rho_G$ correspond to the structural characteristics of the strategic interactions, while $\rho_L$ denotes their dialogical conditions. We refer to the former as global rules, and to the other as local rules. $\sigma$ is a set of integers indicating the number of moves available to each agent, which we term ranks. 
\end{definition}

Before detailing the set of rules $\rho$, we present the notational peculiarities of DDG. The DDG formalism evokes dialogical logics \citep{beirlaen2016inconsistency} and game-theoretic semantics \citep{rahman2000dialogical} to gain expressiveness. In order to facilitate its presentation, and secure a more succinct application, we first introduce the corresponding formalism for a dialogue between two persons (2-DDG).

\begin{definition}[\textbf{2-DDG Formalism}]
    	Let us suppose that players \textbf{P} (the Proponent) and \textbf{O} (the Opponent)  play a DDG having $\psi[\phi_0,\ldots,\phi_{n-1}]$ as its initial thesis, i.e., beginning with the claim that the conclusion $\psi$ follows from the premises $\phi_0,\ldots,\phi_{n-1}$. The players' moves are assertions or requests which may serve to attack ($A$), defend ($D$) or inquire about a particular premise. Moves are represented as expressions of the form \textbf{X-}$e$, where \textbf{X} is either \textbf{P} or \textbf{O}, and $e$ stands for either an assertion, i.e., either $A$ or $D$, or a request, i.e. a query. The symbols \say{!} and \say{?} signal the agents' assertions and requests, respectively. 
	
    	Agents' ranks $r_i\in\mathbb{N}, i\in\{1,2\}$ with \textbf{O}$\to 1$ and \textbf{P}$\to 2$ indicate the number of attacks and defences they can play within the DDG. Agents assert their ranks as so: \textbf{O-}$n:=r_1$ and \textbf{P-}$m:=r_2$. The counters of moves in DDG are denoted by $P_{\mathcal{D}}(\cdot)$, and are also referred to as the game's position\footnote{In a purely semantic context, the game's position is given by subformulas which are derivable from the (syntactical) formulas at play, thus denoting the stage of semantical elaboration of the strategic interaction (see the Hintikka paper)}.
\end{definition}

\begin{definition}[\textbf{Strategy}]
    A strategy of player $\textbf{X}$ in $\D$ is a function $s_x$ that assigns a legal $\textbf{X}$-move to every non-terminal play, the last member of which is a $\textbf{Y}$-move.
\end{definition}{}

\begin{definition}[\textbf{Local 2-DDG Rules ($\rho_L$)}]
    The local rules of a 2-DDG are as follows
    \begin{table}[H]
	\centering
	\begin{tabular}{c|c|c}
		Assertion & Attack & Defence\\\hline
		\textbf{X-}$!\phi\land\psi$ & \textbf{Y-}$?\land_{L}$/\textbf{Y-}$?\land_{R}$ & \textbf{X-}$!\phi$ /\textbf{X-}$!\psi$\\
		\textbf{X-}$!\phi\lor\psi$  & \textbf{Y-}$?\lor$ & \textbf{X-}$!\phi$/ \textbf{X-}$!\psi$\\
		\textbf{X-}$!\neg\phi$	    & \textbf{Y-}$!\phi$ & --\\    		
		\textbf{X-}$!\phi\to\psi$ & \textbf{Y-}$!\phi$  & \textbf{X-}$!\psi$\\    		
		\textbf{X-}$!\psi[\phi_0,\ldots,\phi_{n-1}]$ & \textbf{Y-}$!\phi_0,\ldots,$\textbf{Y-}$\phi_{n-1}$ & \textbf{X-}$!\psi$\\    		    		
	\end{tabular}
	\caption{Local Rules of a Formal Dialogue \citep{beirlaen2016inconsistency}}
	\label{tab:dialog_lrules}
    \end{table}
\end{definition}{}

\begin{definition}[\textbf{Global 2-DDG Rules ($\rho_G$)}]
    The global rules of a 2-DDG are as follows
    \begin{table}[H]
	\centering
	\begin{tabular}{P{0.12\textwidth}|P{0.8\textwidth}}
		\hline
		\textbf{Rule}  &   \textbf{Definition} \\\hline
		Dialogue Starter	&  \specialcell{$P_{\mathcal{D}}(\textbf{P-}!\psi[\phi_0,\ldots,\phi_{n-1}])=0$,\\ $P_{\mathcal{D}}(\textbf{O-}n:=r_1)=1$ and $P_{\mathcal{D}}(\textbf{P-}m:=r_2)=2$}\\\hline
		Move Order  &  \specialcell{$F_{\mathcal{P}}(M):=[m, Z], m<P_{\mathcal{D}}(M), Z\in\{A,D\}$\\ \specialcell{$[\mathcal{P}]_{-1} = \textbf{Y-}e, M_0\equiv[\mathcal{P}]_{0} = \textbf{Y-}e$},\\ there are $n$ $\textbf{X-}e$ moves s. t. $F_{\mathcal{P}}(M_0)=\ldots=F_{\mathcal{P}}(M_{n-1})=[m_0, Z] Z\in\{A,D\}$} and a move $N := \textbf{X-}f$ s. t. $F_{\mathcal{P}\cup\{N\}}(N)=[m_0, Z]$ iff $n<r$ where \textbf{X-}$l:=r$\\\hline
		Classical Attack  &  \specialcell{$N = \textbf{P-}!\psi\in\mathcal{P}$,\\ then there exists $N = \textbf{O-}!\psi\in\mathcal{P}$ s. t. $P_{\mathcal{P}}(M)<P_{\mathcal{P}(N)}$}\\\hline		
		Winning Criterion  & Agent \textbf{X} wins $\mathcal{P}$ iff there is no move $Q=\textbf{Y-}g$ s.t. $\mathcal{P}\cup\{N\}\in\mathcal{D}$ whenever $[\mathcal{P}]_{-1}=\textbf{X-}e$ \\
		\hline
	\end{tabular}
	\caption{Global Rules of a Formal Dialogue}
	\label{tab:dialog_grules}
	\end{table}
\end{definition}

While noting that the winning criterion is semantically equivalent to a relation of entailment, let us also observe that 2-DDG describes a process of semantic inference itself --- that is, it is a type of semantics game \citep{hintikka1997game}. Distinct from classical games, 2-DDG gives a more robust grounding to the strategic interaction occurring between the two players, to the point that the game can be approached as a dialogue. This interpretation extends VAFs in the sense that argumentation frameworks, originally conceived as a backdrop for comparing arguments, become structured environments of dialogue\footnote{As  demonstrated in Afanador(2020), d-models can be seen as VAFs with additional structure. VAF's orders define a d-model's semilattice, whose set of worlds can be identified with the former's set of motivational states.}. To see this more clearly let us consider two minimal examples.

The following examples showcase two situations where a Paramuno engages in a dialogue with an external Researcher, taking on the roles of the Proponent and Opponent, respectively. The dialogue in Example \ref{ex:ddg_1} is centered on argument $A_6$ of Example \ref{ex:main}, posing the contradiction generated by reasoning about agriculture merely as an ESF-mediated economic activity. The dialogue in Example \ref{ex:ddg_2} revolves around a new argument which makes explicit the contradiction ensuing from the interrelation between the use of ecosystem services conveyed in propositions $a$ and $h$.

Both examples are represented in a modified normal form. The numbers in parenthesis indicate the index of the corresponding move, while the numbers on the opposite side specify the index of the competing argument. In square brackets appear alternative moves, i.e, the branches of the game in extended form.

\begin{example}[\textbf{Arguing the Inconsistent Thesis $A_6$ with 2-DDG}]\label{ex:ddg_1}
 Given the contradiction in Example \ref{ex:main} $A_6\iff \{a\land\neg a\}$, a 2-DDG between a Researcher and a Paramuno, respectively, opposing and proposing this argument, has the following form,
 
	\begin{table}[H]
		\centering
		\begin{tabular}{ ccc | ccc }
			\multicolumn{3}{c}{Researcher (\textbf{O})} & \multicolumn{3}{c}{Paramuno (\textbf{P})} \\\hline
			&       &       &       &   \textbf{P-}$!a\land\neg a$   &   (0)\\
			(1)   &   \textbf{O-}$?\land_R$ [\textbf{O-}$?\land_L$]  &   0   &       &       \textbf{P-}$!\neg a\,$[\textbf{P-}$!a$]     &   (2)\\
			\multicolumn{6}{c}{The Paramuno wins}
		\end{tabular}
%		\caption{Arguing an Inconsistent Thesis in LD under $D\langle 1,1\rangle$}
		\label{tab:incons_example_1}
	\end{table}
% 	Aside from the more straightforward rules coming to mind, let's suppose that the Opponent may only use an atomic statement already used by the Proponent. The winning criterion being the responding agent's inability to generate new assertions without producing repetitive.
\end{example}{}

\begin{example}[\textbf{Arguing the Inconsistent Thesis $A_7$ with 2-DDG}]\label{ex:ddg_2}
 Given the contradiction in Example \ref{ex:main} 
 $$\footnotesize A_7:=\langle\{a, a\to \neg h, (\neg h\to\neg s)\to (\neg s\to\neg y)\}, \neg a\rangle\iff \{(a\land\neg a)\to \neg a\},$$ a 2-DDG between a Researcher and a Paramuno, respectively, opposing and proposing this argument, has the following form,
 
    \begin{table}[H]
    \centering
    	\begin{tabular}{ ccc | ccc }
    		\multicolumn{3}{c}{Researcher (\textbf{O})} & \multicolumn{3}{c}{Paramuno (\textbf{P})} \\\hline
    				&       	 		 &       &       &   \textbf{P-}$!(a\land\neg a)\to\neg a$   &   (0)\\
    		(1)   	&  \textbf{O-}$a!\land\neg a$	 &   0   &       &        \textbf{P-}$!\neg a$		&   (2)\\
    		(3)   	&  		\textbf{O-}$!a$			 &   2   &       &       	---		  	&      \\
    		(5)   	&    \textbf{O-}$!\neg a$		 &       &   1   &           \textbf{P-}$?R$   	  	&   (4)\\
    				&    				 &       &   5   &           \textbf{P-}$!a$   	  	&   (6)\\\hline
    		\multicolumn{6}{c}{The Paramuno wins}
    	\end{tabular}
    	%		\caption{Arguing an Inconsistent Thesis in LD under $D\langle 1,1\rangle$}
    	\label{tab:incons_example_2}
    \end{table}
\end{example}{}

In Example \ref{ex:ddg_1}, as the Researcher may only use an atomic statement already used by the Paramuno, responding to the agent's inability to generate new assertions without producing a repetitive strategy modifies 2-DDG's winning criterion. This situation results in the contradiction implicated by argument $A_6$ becoming a viable statement about which the Researcher and the Paramuno can reason. It indicates that the Paramuno predicament of both giving up the cultivation of some Páramo sites while withdrawing from others, does not occur as a consequence of irrational or uninformed behaviour, but rather as a logical consequence of their objective conditions.

Example \ref{ex:ddg_2} further tells us that not only is the contradiction valid, but that refraining from growing crops is also a valid (and non-trivial) consequence of the contradiction. The recognition of the fact that the páramo plays an indirect role in the provision of water used in agriculture, while inscribed in the contradiction expounded on $A_6$, opens up the possibility of opting out of agriculture as a (dialectic) logical consequence, with objective grounding; rather than a mediated determination or ad-hoc imposition.

\subsection{Dialectical Games are not Irrational}

Example \ref{ex:ddg_1} indicates that our approach is paraconsistent, for 2-DDG can operate on inconsistencies. Example \ref{ex:ddg_2} shows that our approach is also dialetheic, meaning that inconsistencies are not only valid but that they can be operated upon as to generate new conclusions, e.g., on distinct ways of using the Páramo. We now proceed to show that DDG, and by implication our approach, is not irrational, in the sense that Arisotle's principle of non-contradiction is correct under its subjacent logic.

Let us, note that DDG's d-model is endowed with the reversal operator $*$, and an order $\leq$ which behaves like cyclic relation. The reversal operator takes a single situation to its dual, e.g., takes $B:=(\neg A \land A)$ to $B^{*}:=\neg\neg A \lor \neg A$ given a well-formed formula $A\in S$. The new order relation, for its part, establishes a terniary order among world situations, i.e., the order $b\leq_a c$ of situations $a,b,c\in W$ indicates that situation $b$ precedes $c$ from the perspective of $a$ or, equivalently, that after $a$, $b$ has to be reached before $c$. Under these definitions, statements in $S$ take on four possible values:

\begin{itemize}
    \item $t$: $a$ holds in $B$ but $\neg a$ does not.
    \item $i$: $a$ and $\neg a$ hold in $B$.
    \item $n$: neither $a$ nor $\neg a$ hold in $B$.
    \item $f$: $a$ does not hold in $A$ but $\neg a$ does.
\end{itemize}{}

Put differently, the logic of contradiction underpinning 2-DDG can be thought of as a 4-valued intentional one over the set $\{t,i,n,f\}$, where $t$ and $f$ represent the classical notions of validity, and $i$ and $n$ account for the possibility of encountering inconsistent or incomplete arguments, respectively. So, by modifying the d-model's (VAF's) $val$ function accordingly, i.e., by updating the function's image to the set $T:=\{t,i,n,f\}$, and introducing a notion of interpretation through a function $I:\A  \times W \mapsto T$ --- where $W$ is a nonempty subset of $\A$, i.e., a set of possible worlds --- we can determine the basic semantic properties of DDG's model of entailment, i.e., DDG's d-model, from the following conditions \citep{anderson2017entailment}:

    \begin{description}
    \item[C1]: $I(p,A) = val(p,A)$ for every proposition $p \in A\subset\A$.
    \item[C2]: $I(a\land b,C)$ iff $I(a,A)=t=I(b,A)$.
    \item[C3]: $I(a\to b,C)=t$ iff, for every $B,C\in W$, $I(b,C)=t$ whenever $I(a,B)=t$ and $b\leq_a c$.
    \item[C4]: $I(\neg a,A)=t$  iff $I(a,A^{*})\neq t$.
    \end{description}

These conditions guarantee that most standard results on truth are forthcoming, and that there are contradictory statements which are simultaneously true and, in consequence, valid. Say, we have a DDG's d-model $\mathcal{M}$ such that for every argument $C$, $I(C,R)=t$ iff $C$, for some $R\in W$, where $W$ is the set of possible worlds of $\mathcal{M}$. Then an argument $A$ is (classically) false (in DDG's d-model) iff $I(A,R)=f$; and truth is the property of all and only those arguments which are true in DDG's d-model $\mathcal{M}$. It is never the case that $I(A,R)=f$ and $I(A,T)=t$, i.e., no argument is both true and false, but every argument is either true or false, and false statements are unprovable. It follows that, since $\neg A(A\land \neg A)$ is a (provable) theorem, it is valid in our d-model, hence true, i.e., DDG's d-model is not an irrational model of entailment.

The importance of abiding by Aristotle's principle, while allowing for contradictions as objective constituents of sociality, consists in the possibility of obviating the Kantian idea --- underpinning value fetishism --- that consistency imposes a bound on rationality \citep{wedin2004use}. Just like Hilbert's idea that consistency ensures mathematical correctness \citep{sieg2012shadow}, the identification of logical investigations with a preordained notion of the world does not hold. To an extent, DDG vindicates the position that dialectical investigations are viable, for Aristotle correctly formulated the principle of contradiction, but erroneously inferred that the real could not be contradictory, and that there are no contradictions in things \citep{routley1976dialectical}. Since capitalist social praxis --- as the substrate of our reality principle --- is contradictory at its core, a paraconsistent and dialetheic analysis is more adequate if the consequences of these contradictions are to be fully grasped.

In the next section, we implement our computational environment with DDG interactions. Our approach operates on the dialogues and interactions captured in Colombia between Paramunos and Researchers, as well as those occurring within the Paramuno community. We reproduce said dialogues to discern distinct forms of appropriating the Páramo, emanating from our paraconsitent and dialetheic analysis of discourse. 
\section{Results}
\label{sec:results}
This section presents the results of running our, now fully fledged, computational environment. The data on which our computational environment operates comes from our interactions with the Paramunos, so we begin by recounting our conversations with them. We then move on to interpreting our results through various graph-theoretic techniques, and advancing some conclusions on how our dialogues with the Paramunos may reshape the Paramo itself, and how they differ from typical VES prescriptions.

Before looking into the practical upshot of our conceptual and analytical ruminations, let us recall that our approach was conceived as in Section \ref{sec:example}. That is, it consists of an ELMO model for processing text data, in natural language, integrated into our DDG model, in the manner of a continuous flow of the arguments and \say{motivational states}  expressed by the people involved in the dialogues (see Figure \ref{fig:av}). Although the NLP model enhances our capacity to work with relatively unmediated dialogical inputs, its role is merely instrumental --- it is the DDG model that is central to our analysis. 

The (text) corpus of our computational environment consists of clauses, processed as arguments and motivational states composed of factual information in the form of premises --- such as \textit{a} (standing for \say{increased agricultural land-use}) or \textit{h} (standing for \say{improved hydrological regulating services}), in Section \ref{sec:example} and Section \ref{sec:logics_contradiction}. Clauses are either facts or rules to derive new information, in the spirit of PROLOG \citep{flach1994simply}. Rules have a head an a body composed of facts, while facts only consist of the head. Rules can be read as the head being true if the body of the rule is found true. Rules can be recursive, i.e., they can be processed by solving smaller instances of the statements they declare, after solving their basic case \citep{flach1994simply}. Clauses indicating motivational states are distinguished from arguments, if they imply a relation of order. This distinction is achieved by calibrating the ELMO model with the contextual information obtained from the transcriptions of informal dialogues with the Paramunos.

\subsection{The Set-up: Conversing with the Paramunos}
Multiple encounters were arranged with the locals. We ran one workshop and engaged in less structured interactions with several Paramunos individually, and as members of a community-based organisation in separate instances. Additional meetings were held with the authorities of the Guanentá-Alto Río Fonce wildlife sanctuary within the GLR complex. Although a research protocol was drafted and observed, some of these encounters occurred fortuitously, enabling an immersive involvement rather than an observational experience.

The workshop was centered on various notions commonly associated with the idea of value. It was divided into two parts, beginning with a discussion on forest management practices as a means to motivating the construction of a time line depicting the various ways in which the Páramo has been appropriated for the past three decades. The second part was devoted to (de)constructing value. It comprised a series of didactic activities revisiting and introducing distinct notions that are, sometimes unknowingly, connected to that of value. Our objective was to prompt a debate on the apparent multiplicity of values, and the existence of a common basis --- if any; for if a type of value is recognised as such, there must be a fundamental quality that it shares with other types of values upholding said recognition. The exploration of these ideas would either corroborate, or not, the extent to which the Paramuno sociality could be considered of a value-based nature.

\subsection{Results from Dialoguing through 2-DDGs}

\begin{figure}
	\centering
	\begin{minipage}{0.5\textwidth}
		\centering	
		\includegraphics[width=\textwidth]{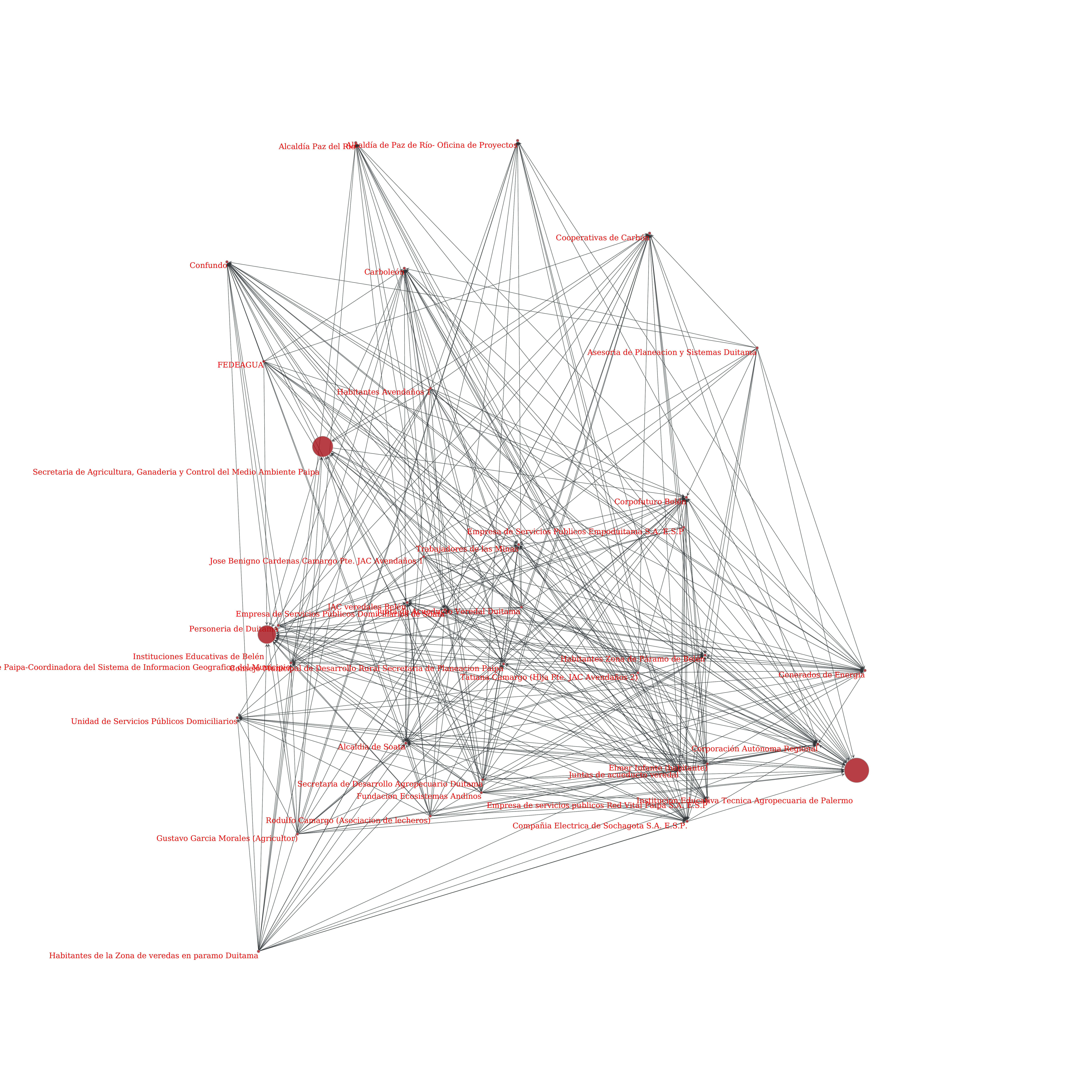}\\ 
    	\caption{GLR Betweenness Network}
    	\label{fig:bet}
	\end{minipage}
	\hspace{-0.5cm}
	\begin{minipage}{0.5\textwidth}
	\centering
		\includegraphics[width=\textwidth]{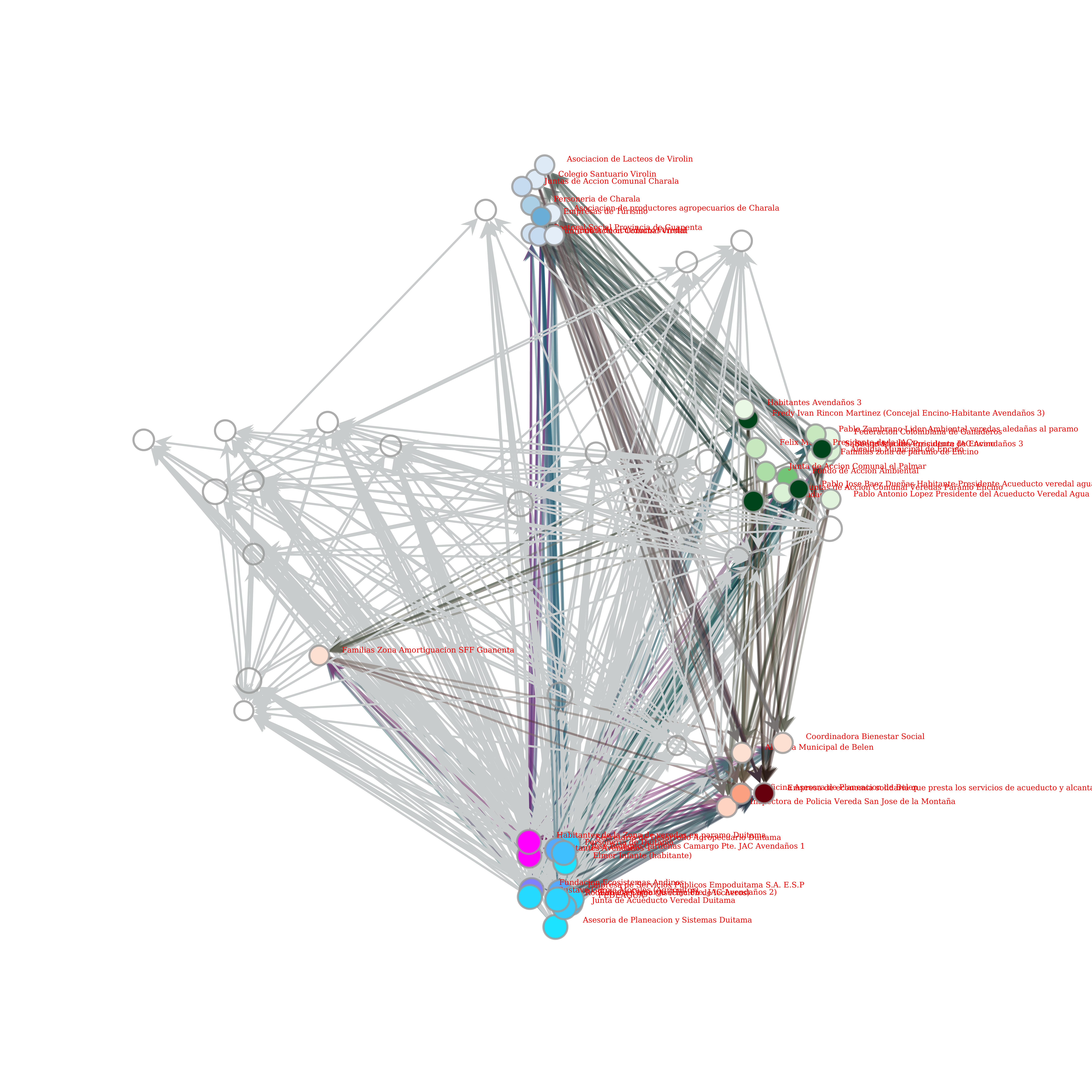}\\ 
    	\caption{GLR Argumentative Network}
    	\label{fig:space_bet}  
	\end{minipage}
\end{figure}

\enlargethispage{1\baselineskip}
We use networks and graph-theoretic techniques to represent our dialogues and interactions with the Paramunos and among themselves. These devices are apt for presenting relational informational, like the corpus at our disposal \citep{newman2001fast}. Informally, a network is a graph with contextual information, i.e., a collection of vertices and weighted edges, indicating certain relational properties. 

Figure \ref{fig:bet} shows the links among several social actors in the Páramo, in accord with the baseline information reported by \cite{rojas2017caracterizacion}. The edges express a common actor typology, and the size of the nodes reflect their degree of betweenness, with respect to the information on water and agropastoral practices available to each actor. Unsurprisingly, grassroots and community led organisations display larger values of betweennes, for they are well-connected, and in a better position to facilitate the transmission of information. Let us recall that betweenness is a centrality measure informing the number of shortest paths in a network that pass through a particular node \citep{opsahl2010node}, as an indication of how much information is likely to flow through said node, i.e., how much first-hand information a Páramo actor is capable of communicating. 

\begin{figure}[h]
	\centering
	\includegraphics[width=.6\textwidth]{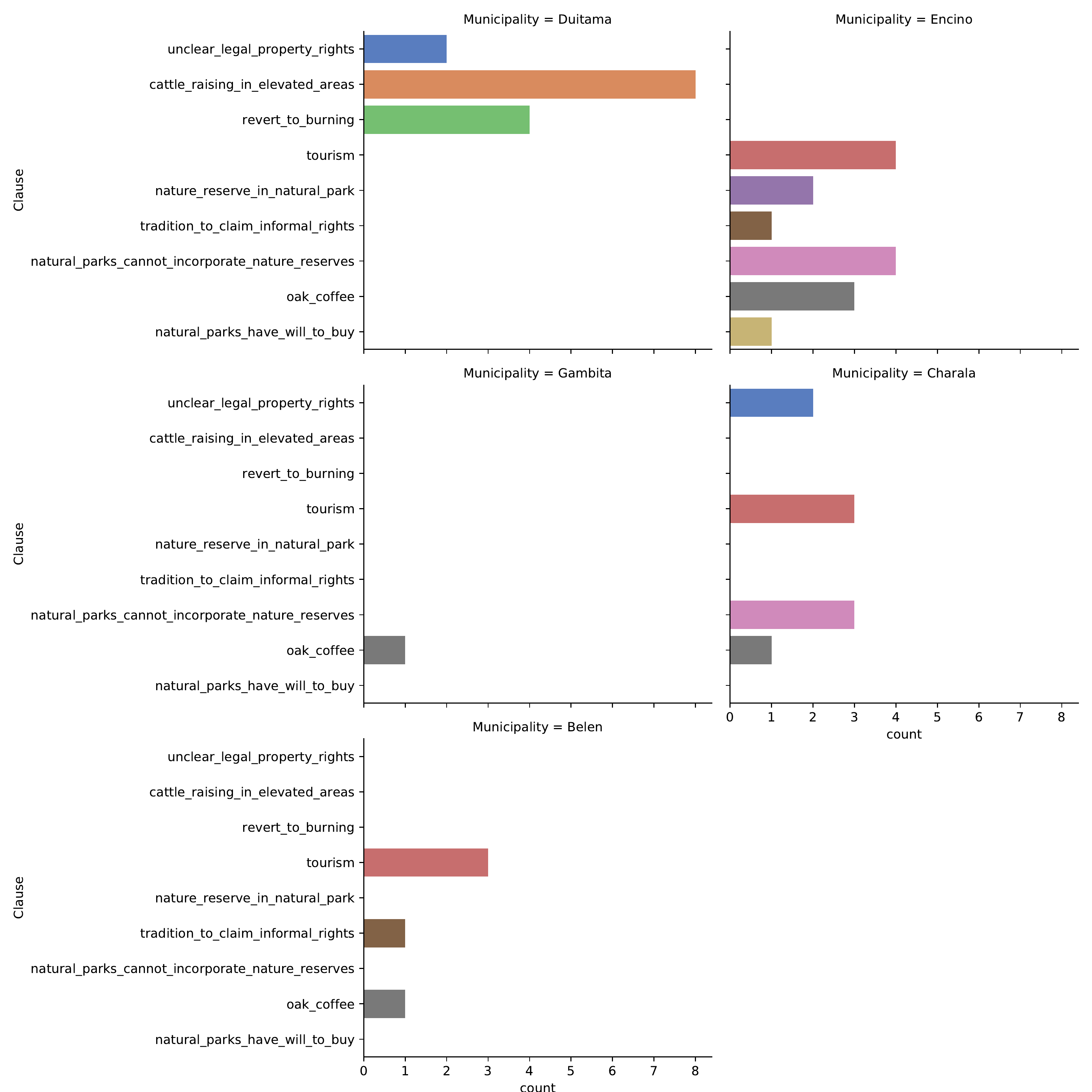}\\ 
	\caption{Identification and Distribution of Premises over the GLR Complex}
	\label{fig:mun_catplot}  
\end{figure}

Nine recurrent premises were identified by ELMO, while processing our corpus: \say{cattle  raising  in  elevated  areas},  \say{natural  parks  cannot  incorporate  nature  reserves},  \say{natural  parks  have  willingness  to  buy},  \say{nature  reserve  in  natural  park},  \say{oak  coffee},  \say{revert  to  burning},  \say{tourism},  \say{tradition  to  claim  informal  rights},  \say{unclear  legal  property  rights}. They are distributed within the Páramo as shown in Figure \ref{fig:mun_catplot}.

Figure \ref{fig:mun_catplot} gives further evidence of the issues raised in Section \ref{sec:paramo}. As the more elevated areas are located within Duitama, the handful of Paramunos who employ traditional practices, involving cattle raising and burning, can be found in this municipality. The (premises making up the) arguments put forward reflect not only this situation, but the ostensibly informal pattern of land occupation.

Unlike Duitama Paramunos, those based in Encino and Charalá, i.e., within the cloud forest belt or lower areas, express a different bond with the Páramo, for they experience the páramo at one remove. Drawn by the pull of the economic drive of Santander, the páramo is seen as a secondary crossing point into Boyacá, a source of productivity and a referent for the construction of social identity. The particularity being that, the semiotic burdens of these two latter meanings are conflated into premises upholding touristic activities, entrepenurial initiatives --- such as the production of \say{oak coffee} from Quercus acorns (see Section \ref{sec:paramo}) --- and the creation of private natural reserves.

\begin{figure}
    \centering
    \includegraphics[width=\textwidth]{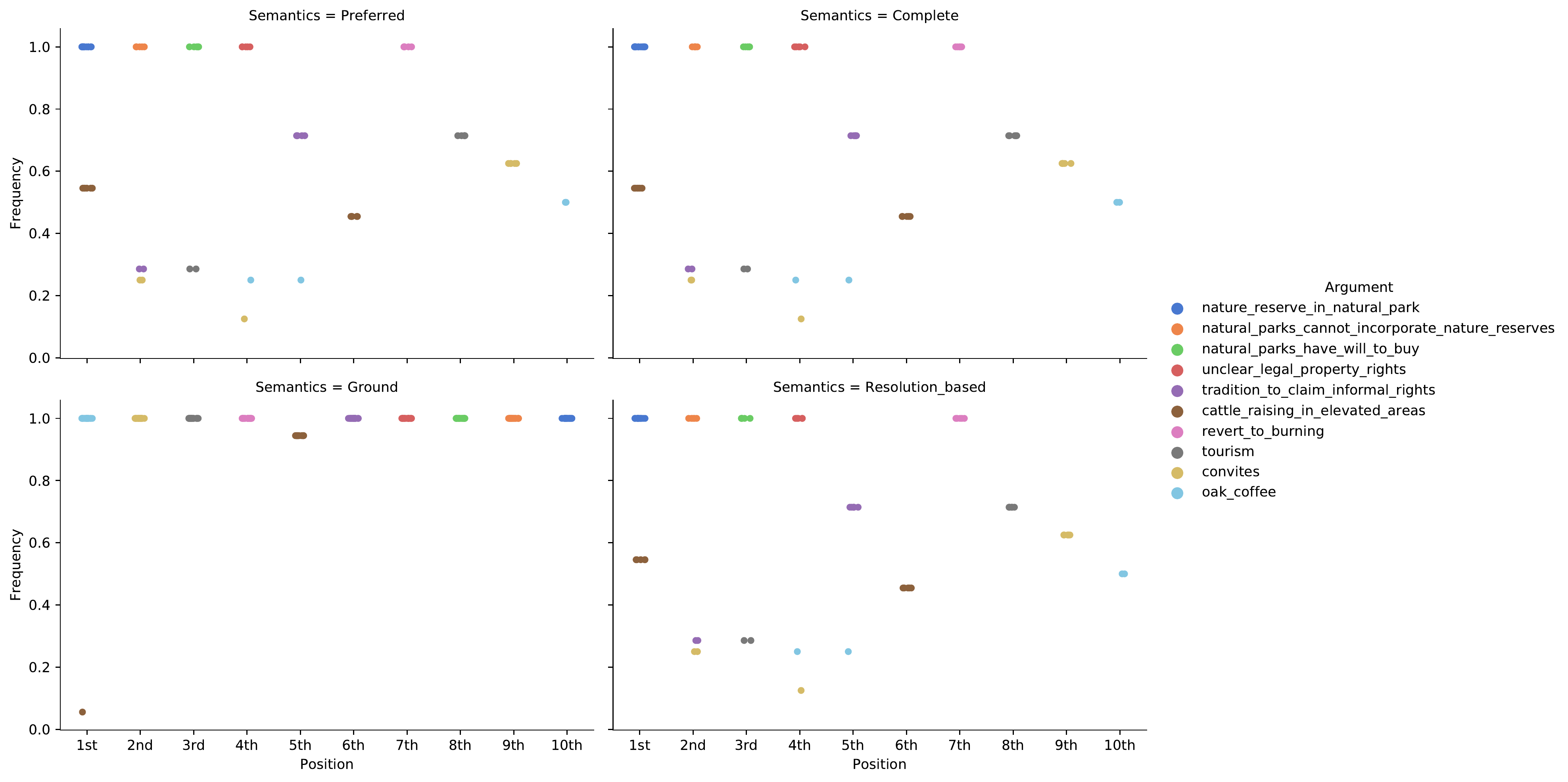}
    \caption{Argument Position over Multiple DDG Semantics}
    \label{fig:semantics}
\end{figure}{}

The graph in Figure \ref{fig:space_bet} displays the same network, but with its nodes geolocated and adjusted by the occurrence of dialogues. Only those nodes corresponding to the persons or institutions taking part in the dialogues appear labelled, and their connecting edges coloured. That is, the dialogues registered in our corpus took place in Charalá, Gámbita, Duitama, Belén, and Encino, the majority of which occurred on the eastern side of the complex, where the bulk of the regional economic activity is concentrated. 

Each plot in Figure \ref{fig:semantics} shows how often the premise of an argument denoting an action on the Páramo appears at a particular position in the final prioritisation orders. These orders denote various arrangements of arguments satisfying the dialogical interactions expressed in 2-DDGs. However, the relative position of each argument does not necessarily represent a preference relation, since the ciclicity of our d-model's orders --- i.e., the formal relations of order within DDG's logical model entailment --- may prevent the construction of a choice function satisfying the basic axioms of (economically rational) preference formation \citep{sen1993internal,hausman2011preference}. Put differently, the orders generated through DDG can be more adequately interpreted as concrete (objective) choices within inconsistent domains, rather than preferences describing mental states complying with (classically logical) rational behaviour \citep{gul2008case}.  

The semantics of acceptance applied to DDG are based on the argumentation-theoretic concept of extension. The results correspond to preferred, complete, grounded and resolution-based grounded extensions (see Section \ref{sec:example}). A set of arguments in a DDG is \textit{complete} if 
its members are conflict-free in the sense that they do not attack one another, and if they have all survived dialogues similar to Example \ref{ex:ddg_2} \citep{besnard2009argumentation}. A minimal (with respect to set-theoretic inclusion) complete set of arguments constitutes a \textit{grounded} extension in DDG \citep{cerutti2014}. Its maximal (with respect to set-theoretic inclusion) counterpart is termed \textit{preferred} extension \citep{Walton2009}. A preferred extension that is (topologically) closed is further qualified as a \textit{resolution-based} extension \citep{baroni2011resolution}.

Complete, preferred and resolution-based extensions are, thus, comprised of relatively less stable arguments. Grounded extensions, on the contrary, impose stringent conditions on the arguments surviving DDGs (see the plot at the lower left of Figure \ref{fig:semantics}), which makes them robust to dialogical attacks. After reproducing existing --- and simulating new --- dialogues, while observing the topology of our network in Figure \ref{fig:space_bet} and DDG's semantic properties, we conclude that: 

\begin{enumerate}
    \item The highest ranked actions --- i.e.,\say{natural  parks  cannot  incorporate  nature  reserves},  \say{natural  parks  have  willingness  to  buy},  \say{nature  reserve  in  natural  park}, \say{unclear  legal  property  rights}, and \say{convites} --- involve multiple inconsistencies, residing in the legal and formal obstacles to extend protected areas within the Páramo through the incorporation of private nature reserves.
    \item Reverting to burning and cattle grazing at elevated locations is a likely outcome --- i.e., these actions occur at the first position of the prioritisation order in one out of every two dialogues --- whenever, e.g., inconsistencies are probed for a resolution, as demonstrated by the non-grounded extensions in Figure \ref{fig:semantics}.
    \item The application of DDGs with grounded semantics (see the plot at the lower left of Figure \ref{fig:semantics}) reveals that tourism and entrepreneurial activities, aligned with conservation objectives, induce an order where inconsistencies are appeased, in the sense that the fact that \say{natural  parks  cannot  incorporate  nature  reserves}, on account of \say{unclear  legal  property  rights}, has to be dealt with before addressing those arguments advocating the creation of \say{nature  reserve(s) (with)in (the) natural  park}.
    \item \textit{Convites} --- i.e., traditional forms of cooperative and reciprocal labour associated with the Paramuno identity as much as burning and grazing \citep{rivera2011guia} --- appear in the grounded extension as one of the most robust arguments, only second to the oak acorn drink initiative, suggesting that alternative forms of social praxis offer a way of coping with the contradictions engendered by value-based relations. 
\end{enumerate}{}

\subsection{Network-wide Implications}

\begin{figure}
	\centering
	\begin{minipage}{0.5\textwidth}
		\centering	
		\includegraphics[width=\textwidth]{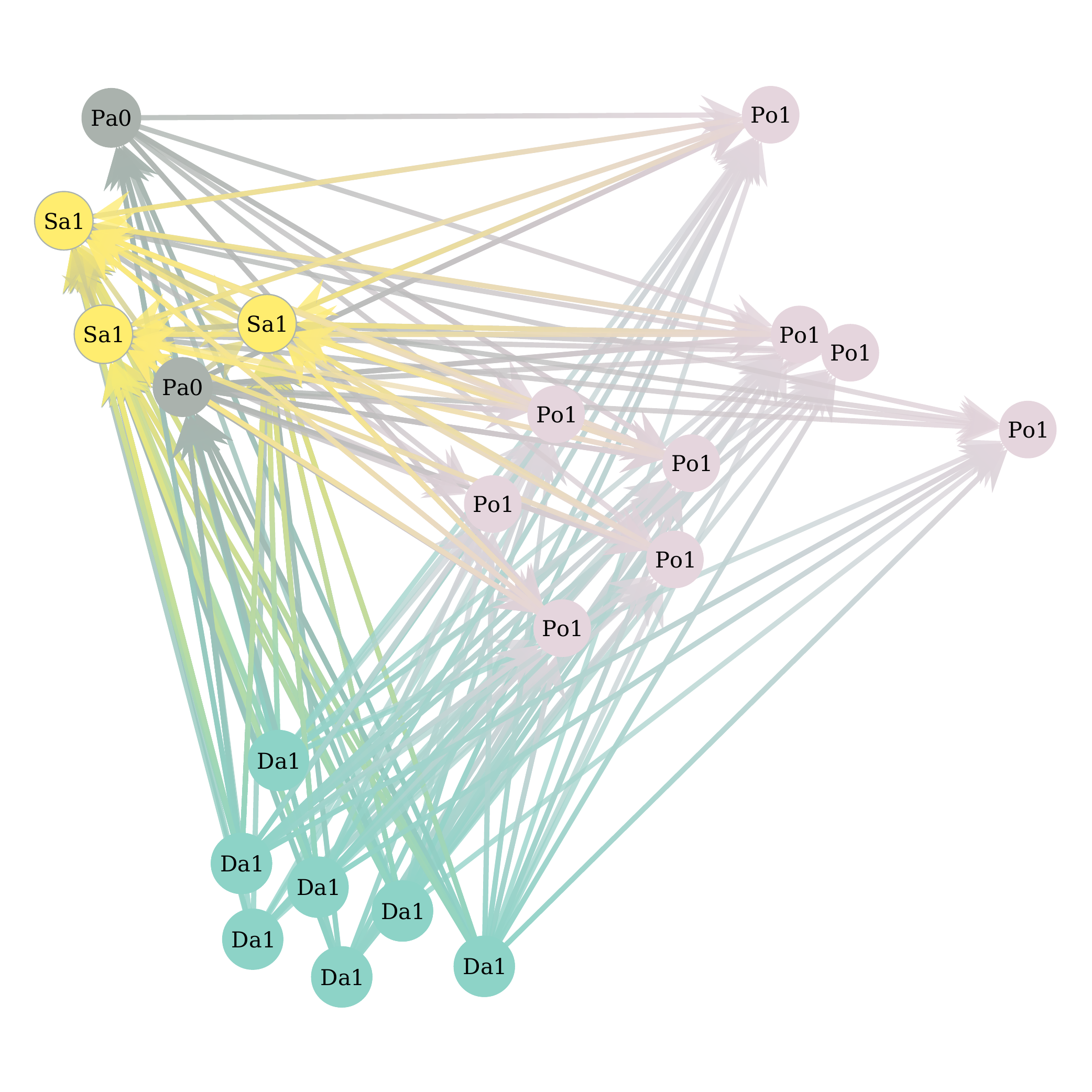}\\ 
    	\caption[caption]{Stable Block Configuration of the \\\hspace{\textwidth} Original GLR Network}
    	\label{fig:bgraph}
	\end{minipage}
	\hspace{-0.5cm}
	\begin{minipage}{0.5\textwidth}
	\centering
		\includegraphics[width=\textwidth]{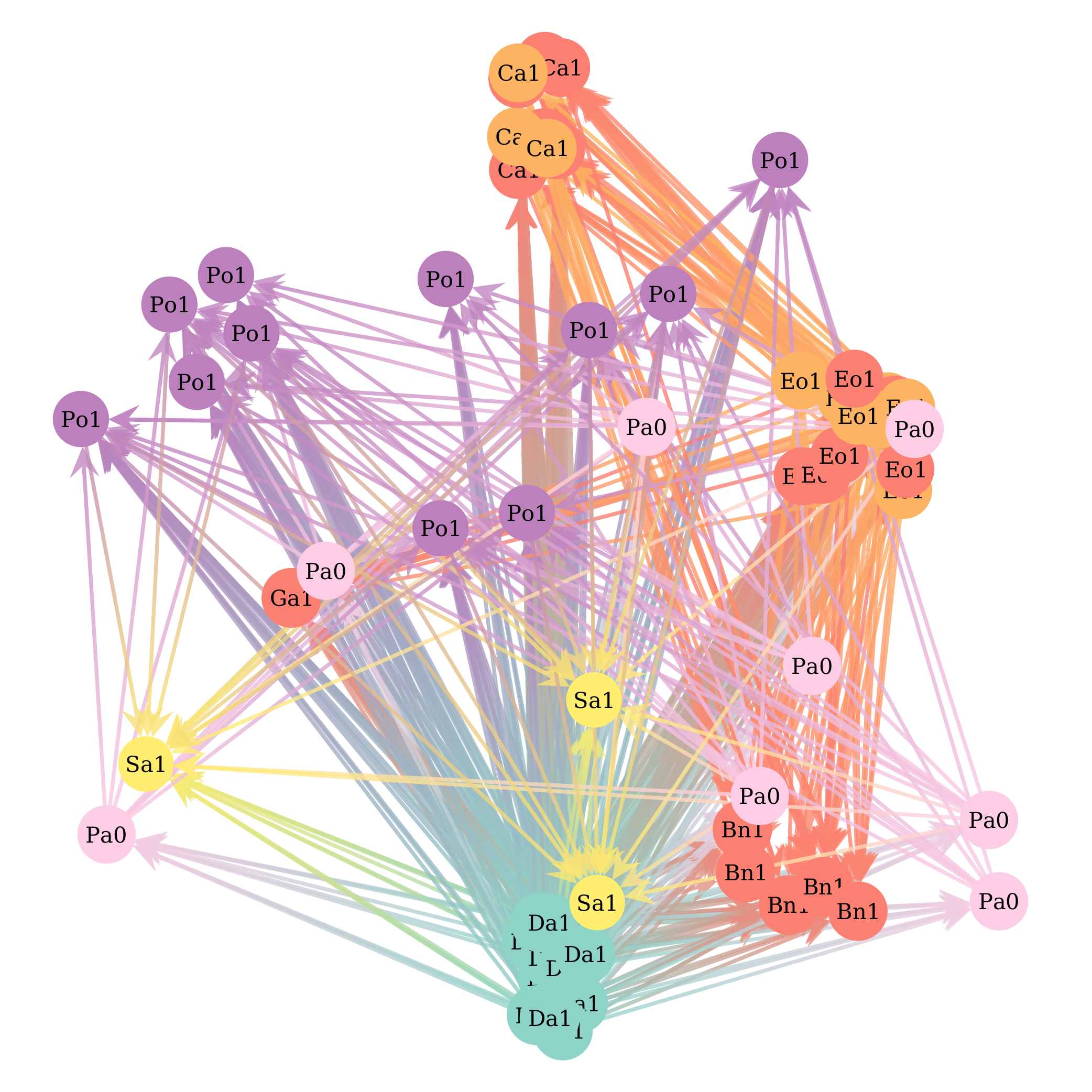}\\ 
    	\caption[caption]{Stable Block Configuration of the \\\hspace{\textwidth} GLR Network under Grounded DDG}
    	\label{fig:semantic_bgraph}  
	\end{minipage}
\end{figure}

Rather than conducting a hypothesis test on the prioritisation data, we situate the statistical evaluation of our findings directly in the analysis of the GLR network. The evaluation of our results involves estimating the stable state of the (argument-based) GLR network's topology, as described by the dialogical outcomes of grounded DDGs. We do this to foretell the likelihood of long-term changes in the configuration of the existing social relations in the Páramo, attributable to the actions suggested by our approach.  

The graphs in Figure \ref{fig:bgraph} and Figure \ref{fig:semantic_bgraph} represent stable configurations of the networks in Figure \ref{fig:bet} and Figure \ref{fig:space_bet}, respectively. The new graphs were obtained by minimizing the microcanonical entropy of the different network realisations (see \cite{bianconi2009}) enabled by grounded DDG interactions, through an agglomerative MCMC heuristic \citep{peixoto2013}. This means that the entropy induced by our grounded dialogues is used to estimate the probability of nodes and edges --- i.e., Páramo actors and their relations --- of falling within a particular block partition of the GLR network, by way of a(n) (algorithmically) greedy MCMC sampling of the blocks --- i.e., an agglomerative heuristic \citep{peixoto2014}.

\begin{table}
    \centering
    \subfloat[Stable Baseline Network]{
    \begin{tabular}{lrr} 
		\toprule 
		Municiplaity &  Block &  Occurrences \\ 
		\midrule      Duitama &      0 &           14 \\        Paipa &      1 &            8 \\  Paz del Rio &      2 &           11 \\        Soata &      3 &            3 \\ 
	    \bottomrule 
	\end{tabular}
	}
  	\hfill
  	\subfloat[Stable Network under Grounded DDG]{
 	\begin{tabular}{lrr} \toprule 
    	Municiplaity &  Block &  Occurrences \\ 
    	\midrule 
         Belen &      4 &            6 \\      Charala &      2 &            4 \\      Charala &      4 &            7 \\      Duitama &      0 &           12 \\       Encino &      2 &            9 \\       Encino &      4 &            8 \\      Gambita &      4 &            1 \\        Paipa &      5 &            7 \\  Paz del Rio &      3 &           9 \\        
         Soata &      1 &            3 \\
    	 \bottomrule
  	\end{tabular}
  	}
  	\caption{Distribution of Actors within Block Configurations}
    \label{tab:blocks}
\end{table}{}

\begin{table}
    \centering
    \resizebox{12cm}{4cm}{
    \subfloat[Stable Baseline Network]{
	\begin{tabular}{lrlr}
		\toprule
		Municiplaity &  Block &       Typology &  Occurrences  \\ 
		\midrule      
		Duitama &      0 &   Conservation &            1  \\      Duitama &      0 &  Institutional &            5  \\      Duitama &      0 &      Productive &            2  \\      Duitama &      0 &         Social &            6  \\        Paipa &      1 &      Academic &            1  \\        Paipa &      1 &  Institutional &            4  \\        Paipa &      1 &      Productive &            2  \\        Paipa &      1 &         Social &            1  \\  Paz del Rio &      2 &      Academic &            1  \\  Paz del Rio &      2 &   Conservation &            2  \\  Paz del Rio &      2 &  Institutional &            3  \\  Paz del Rio &      2 &      Productive &            3  \\  Paz del Rio &      2 &         Social &            2  \\        Soata &      3 &  Institutional &            3  \\ 
		\bottomrule 
	\end{tabular}
	}
	\hspace{3 cm}
	\subfloat[Stable Network under Grounded DDG]{
    \begin{tabular}{lrlr} 
        \toprule 
		Municiplaity &  Block &       Typology &  Occurrences \\
		\midrule   
		Belen &      4 &  Institutional &            5 \\      Charala &      2 &      Productive &            2 \\      Charala &      2 &         Social &            1 \\      Charala &      4 &      Academic &            1 \\      Charala &      4 &  Institutional &            2 \\      Charala &      4 &      Productive &            1 \\      Charala &      4 &         Social &            2 \\      Duitama &      0 &   Conservation &            1 \\      Duitama &      0 &  Institutional &            5 \\      Duitama &      0 &      Productive &            2 \\      Duitama &      0 &         Social &            6 \\       Encino &      2 &   Conservation &            1 \\       Encino &      2 &  Institutional &            1 \\       Encino &      2 &      Productive &            1 \\       Encino &      2 &         Social &            5 \\       Encino &      4 &      Academic &            1 \\       Encino &      4 &  Institutional &            1 \\       Encino &      4 &         Social &            5 \\      Gambita &      4 &      Productive &            1 \\        Paipa &      5 &      Academic &            1 \\        Paipa &      5 &  Institutional &            4 \\        Paipa &      5 &      Productive &            2 \\        Paipa &      5 &         Social &            1 \\ 
		Paz del Rio &      3 &      Academic &            1 \\  
		Paz del Rio &      3 &   Conservation &            2 \\  
		Paz del Rio &      3 &  Institutional &            3 \\  
		Paz del Rio &      3 &      Productive &            3 \\  
		Paz del Rio &      3 &         Social &            2 \\   Soata &      1 &  Institutional &            3 \\ 
		\bottomrule 
	\end{tabular}
	}
	}
    \caption{Block Composition by Actor Typology and Municipality}
    \label{tab:b_typology}
\end{table}{}

\begin{figure}
	\centering
	\subfloat[Stable Baseline Network]{
	\includegraphics[width=.4\textwidth]{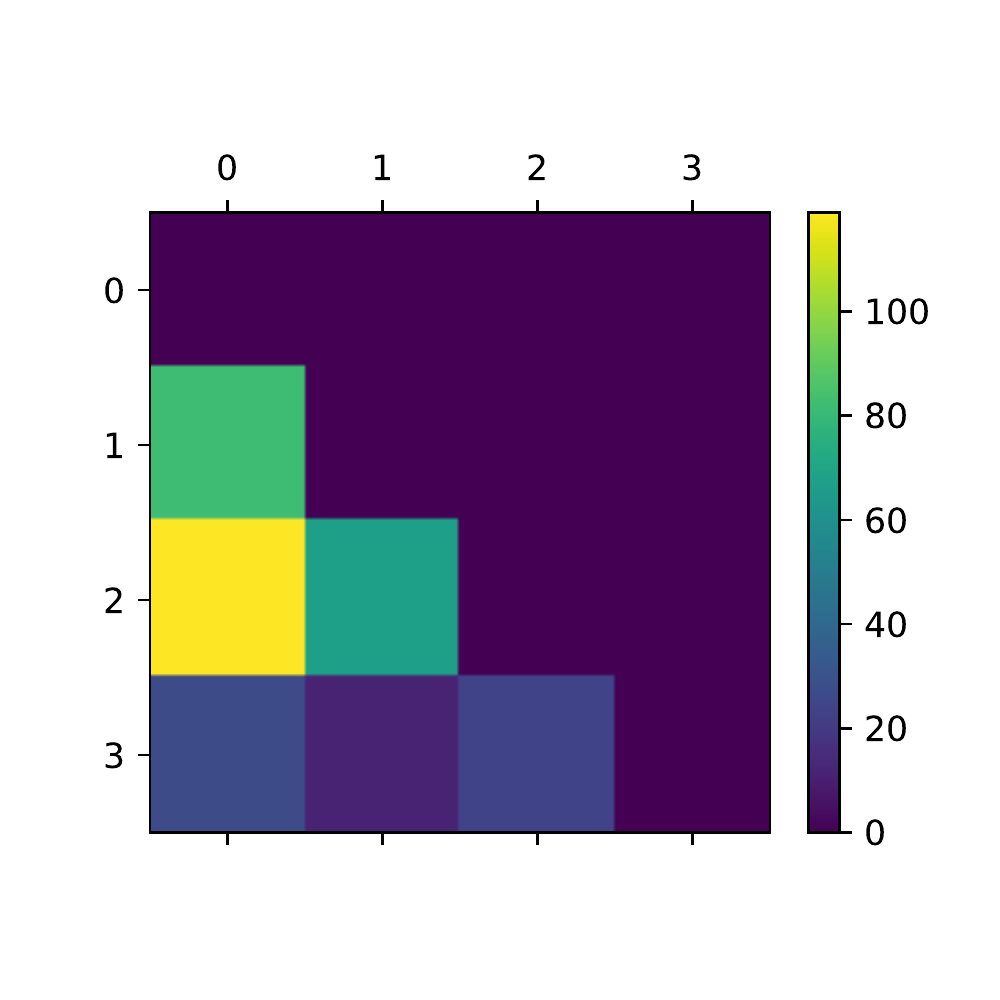} 
	}
	\hfill
	\subfloat[Stable Network under Grounded DDG]{
	\includegraphics[width=.4\textwidth]{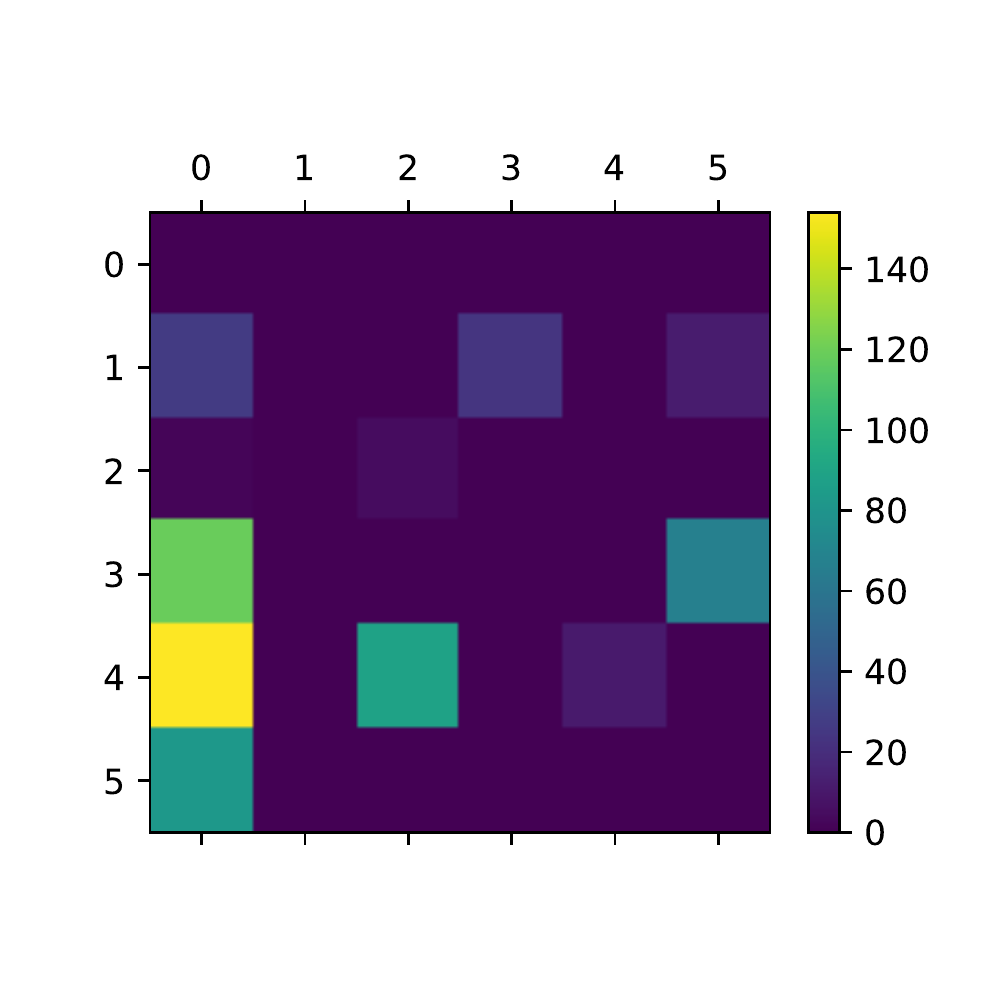} 
	}
    \caption{Block Correlations}
    \label{fig:b_cor}
\end{figure}

Figure \ref{fig:bet} indicates that the municipalities of Duitama, Paipa, Paz del Río and Soata, originally, make up the four blocks explaining the socio-economic dynamics of the GLR complex. Duitama and Paz del Río concentrate most of the actors belonging to the productive and institutional typologies described in \cite{rojas2017caracterizacion} (see Table \ref{tab:blocks} (a) and Table \ref{tab:b_typology} (a)). These actors were relatively unconcerned with the grassroots processes occurring on the eastern parts of the complex, in a scenario prior to the official delimitation of the páramo \citep{rivera2011guia, rojas2017caracterizacion}.

Figure \ref{fig:semantic_bgraph}, shows how that the original network shifts into a six block configuration, once our results are accounted for. Under these new circumstances, some of the actors in Duitama and Paipa loose their importance, judged in terms of block-connectivity, whereas the social actors of municipalities like Encino and Charalá --- namely, community-based associations and the Natural Parks authority --- gain notoriety. It is also worth noting that the municipalities of Charalá and Gámbita establish new bonds across the geographical boundaries of the GLR complex. 

From South to North, Duitama is now connected to Charalá, by way of the integrative efforts of grassroots associations mobilising (geographically) disconnected Paramunos towards collective projects revolving around tourism. From West to East, Gámbita appears as pivotal to consolidate the conservation initiatives put forward by the authorities based in Encino, as the Gámbita Paramunos have devoted large areas to the conservation of Quercus forests and still rely on convites and similar practices. Table \ref{tab:b_typology} details the blocks by municipalities and actor typology.

\newpage

Further to the changes in the spatial connectivity of the Paramunos, evidenced through Figures \ref{fig:bgraph} -- \ref{fig:semantic_bgraph} and Tables \ref{tab:blocks} -- \ref{tab:b_typology}, we examined the qualitative connections highlighted by our DDG analysis. To this effect we estimated the correlation between the blocks in each configuration, using the information on agropastoral and water management practices reported in \cite{rojas2017caracterizacion}, and our previous results. Let us note in passing, that the ensuing analysis does not necessarily conform to a spatial interpretation, for the various configurations of our networks of agents were obtained by looking at criteria which are not directly dependent on geographical considerations --- e.g., block 4 of the grounded network is composed of Páramo actors from the Belén, Charalá, and Encino municipalities (see Table \ref{tab:blocks} (b)).

Figures \ref{fig:b_cor} (a) -- (b) show the degree of correlation among the blocks of our baseline and grounded networks, measured as the number of edges (links) generated among the corresponding blocks throughout their MCMC sampling. The former figure indicates that all municipalities --- as the blocks of the baseline network can be mapped to municipalities --- are correlated to Duitama, although the dependence is not reciprocal. The unilateral connection between Paz del Río and Duitama --- signalled by the yellow box in Figure \ref{fig:b_cor} (a) --- can be explained by the functioning of one of the largest steelworks in the former municipality, which created a considerable demand for agricultural produce satisfied via Duitama, also serving as a link for the transportation of steel into Bogotá \cite{rojas2017caracterizacion}. To this day Duitama is still considered one of the main commercial and manufacturing centres of the Boyacá province. 

\enlargethispage{1\baselineskip}
While confirming the preponderance of Duitama --- a strong bond with Paz del Río is evidenced by the light green square connecting blocks 3 and 0 --- a series of new relations with the Belén, Charalá, and Encino Paramunos --- all belonging to block 4 --- become apparent in Figure \ref{fig:b_cor} (b). At the same time, the close ties between the National Parks authorities and the community-based associations based in Charalá are now clearly discernible by the formation of an average of 100 links between Páramo actors in blocks 2 and 4. This statistical evidence indicates that the network of relational connections among the Paramunos is resilient from an information-theoretic point of view.  

\section{Conclusion}
\label{sec:conclusion}

In this paper we discern a methodological praxis to cater for the appropriation of the Colombian Páramo, by way of negating the notion of value and related categories. Contra the valuation of ecosystem services (VES), and the ecosystem services framework (ESF), we argue that a historically and materially situated --- hence objective --- approach to value is the better means to grapple with the sociality of the Páramo (see Section \ref{sec:paramo}). By observing the historical specificity of value, we recognise the contradictions it fosters and their effects on the ecology of the Páramo; and by operating on these contradictions, rather than treating or attempting to resolve them, we elucidate more sensible decisions as to how the Páramo can be used and re-signified.

VES is built upon the idea that value is either a subjective notion reflecting a fundamental property of all things in nature, or an ontological category itself. In this way, the social mediating quality of value is lost to subjectifying or essentialising exegeses of our relation with nature. VES, and ESF by implication, are, thus, touted necessary for the incorporation of nature's value into the general process of valorisation underpinning economic growth, so that nature can be better appreciated and conserved. It is as though VES practitioners would set out to solve the problems created by value-based social relations, precisely, by furthering the relations themselves.

The people of the Páramo, i.e., Paramunos, are receptive to the VES ethos. Our time with the Paramunos of Guantiva-La Rusia revealed that they are committed to transitioning from traditional agropastoral practices towards entreprenurial activities centered on agro/eco-tourism, while striving to reinterpret (preserve) campesinos as local subjects of the (world) market. This conscious effort to participate in the (global) dynamics of value reproduction, may provoke the cessation of burning and grazing in the páramo grasslands, but also the adoption of value-based relations which seem to occur by fiat.

Since the Páramo sociality is, in this sense, fraught with contradictions, and also far from trivial, our approach should be amenable to inconsistencies, while being capable of generating meaningful conclusions therefrom, i.e., our approach should be paraconsistent and dialetheic. In consequence, we propose a collection of computational techniques that (re)produce dialogical settings where actions on the Páramo are, intentionally or otherwise, conflated with the notion of value. These techniques delineate an environ for all interested stakeholders to put forward their views on how the Páramo can be construed, and pit these arguments against one another as a means to prioritise over multiple, and often conflicting, management practices. 

At the heart of our computational environment lies a paraconsistent and dialetheic model of logical entailment. Our main contribution is posed in terms of the formalisation and implementation of this model as a new type game-theoretic object --- a Dialectical Dialogical Game (DDG). For DDG emanates directly from our critique of value, it is considered apt for understanding value-related categories, and inferring actions from them. 

Instrumental to the application of DDGs is the natural language processing algorithm --- an ELMO model --- generating the arguments and \say{motivational states} expressed by those involved in the dialogues. Thus implemented, DDGs were used to recreate our interactions with the locals, augmenting the baseline information on the Guantiva-La Rusia páramo reported in \cite{rojas2017caracterizacion}. These latter information also served to delineate a baseline network of Páramo actors upon which to run our dialogues.

The analysis of the resulting networks indicate that:

\begin{enumerate}
    \item The highest ranked actions --- i.e.,\say{natural  parks  cannot  incorporate  nature  reserves},  \say{natural  parks  have  willingness  to  buy},  \say{nature  reserve  in  natural  park}, \say{unclear  legal  property  rights}, and \say{convites} --- involve multiple inconsistencies, residing in the legal and formal obstacles to extend protected areas within the Páramo through the incorporation of private nature reserves.
    \item Reverting to burning and cattle grazing at elevated locations is a likely outcome --- i.e., these actions occur at the first position of the prioritisation order in one out of every two dialogues --- whenever, e.g., inconsistencies are probed for a resolution, as demonstrated by non-grounded solutions of DDG.
    \item The application of DDGs with grounded semantics reveals that tourism, and entrepreneurial activities of the like, induce an order where inconsistencies are appeased, in the sense that the fact that \say{natural  parks  cannot  incorporate  nature  reserves}, on account of \say{unclear  legal  property  rights}, has to be dealt with before addressing any argument advocating the creation of \say{nature  reserve(s) (with)in (the) natural  park}.
    \item \textit{Convites} --- i.e., traditional forms of cooperative and reciprocal labour associated with the Paramuno identity as much as burning and grazing--- appear in the grounded extensions of DDGs as one of the most robust arguments, only second to the oak acorn drink initiative, suggesting that alternative forms of social praxis offer a way of coping with the contradictions engendered by value-based relations. 
    \item From South to North, Duitama and Charalá municipalities are connected, by way of the integrative efforts of grassroots associations mobilising (geographically) disconnected Paramunos towards collective projects revolving around tourism. From West to East, the Gámbita municipality appears as pivotal to consolidate the conservation initiatives put forward by the authorities based in Encino, as the Gámbita Paramunos have devoted large areas to the conservation of Quercus forests and still rely on convites and similar practices.
    \item While confirming the preponderance of Duitama, a series of new relations among the Belén, Charalá, and Encino Paramunos become apparent. Furthermore the close ties between the National Parks authorities and the community-based associations based in Charalá are clearly discernible. The statistical evidence supporting these findings suggest that the network of relational connections among the Paramunos is resilient, from an information-theoretic point of view.  
\end{enumerate}{}

We leave the formal proofs of DDG's properties, and the development of the theory of the greater class of dialetheic semantics games to which it belongs, for future work. Extensions of the simple sentential version of the 4-valued paraconsistent logic, underpinning DDG, is also left for future research. The fruition of this forthcoming work will be decisive to the analysis of an augmenting corpus of dialogues, and the prospect of including more intricate interactions. Finally, it is equally relevant to determine the feasibility of tertiarising the economy of the Paramo --- along the lines of the entrepreneurial initiatives identified through 2-DDGs --- in the face of the Covid-19 crisis. We believe that DDGs, and their accompanying conceptual and methodological frameworks, are apt to conduct this latter task.

\newpage
\bibliographystyle{apalike}
\bibliography{ref}

\end{document}